\documentclass[preprint,12pt, authoryear]{elsarticle}
\usepackage[utf8]{inputenc}
\usepackage[T1]{fontenc}
\usepackage{mathptmx}
\usepackage{amsmath}
\usepackage{amssymb}
\usepackage{calc}
\usepackage[english]{babel}
\usepackage[authoryear]{natbib}
\usepackage[a4paper, lmargin=0.1666\paperwidth, rmargin=0.1666\paperwidth, tmargin=0.1111\paperheight, bmargin=0.1111\paperheight]{geometry}
\usepackage[pdftex,linkcolor=blue,urlcolor=blue,citecolor=blue,pdfborder={0 0 0}]{hyperref}
\usepackage{graphicx}
\usepackage{longtable}
\usepackage{float} 
\usepackage{multirow}
\usepackage{booktabs}
\usepackage{subcaption}
\usepackage{pifont}
\usepackage{threeparttable}
\usepackage[all]{nowidow}
\usepackage[protrusion=true,expansion=true]{microtype}
\usepackage{array}

\journal{}

\makeatletter
\def\ps@pprintTitle{%
  \let\@oddhead\@empty
  \let\@evenhead\@empty
  \def\@oddfoot{\centerline{\thepage}}%
  \let\@evenfoot\@oddfoot}
\makeatother

\begin{document}

\begin{frontmatter}

\title{Examining the Associations between Visual and Non-Visual Elements and Cyclists’ Route Choices for Various Trip Purposes}

\author[doa]{Heyang Hua}
\author[doa]{Koichi Ito}
\author[doa,dre]{Filip Biljecki\corref{cor1}}

\cortext[cor1]{Corresponding author. Email: filip@nus.edu.sg}
\affiliation[doa]{organization={Department of Architecture, National University of Singapore}, 
            addressline={8 Architecture Drive}, 
            city={Singapore},
            postcode={117356}, 
            country={Singapore}}
\affiliation[dre]{organization={Department of Real Estate, National University of Singapore},
            addressline={15 Kent Ridge Drive}, 
            city={Singapore},
            postcode={119245}, 
            country={Singapore}} 

\begin{abstract}
Understanding cyclist preferences for the characteristics of the built environment is important in promoting sustainable urban transportation and active mobility. Despite previous studies on cyclists' route choices, the influence of visual and non-visual factors on these choices for different trip purposes remains unclear; thus, this paper fills this gap through a data-driven case study in Montreal, Canada. Non-visual factors include socioeconomic factors and two-dimensional environments, while visual factors involve visual perception during cycling and are computed using street view images. The study consists of two parts: one part analyzes spatiotemporal information to explore the non-visual factors between the start and end points of cycling trips, and the other part investigates the discrepancies in distributions of these factors between the shortest path and the actual one. The findings reveal the spatiotemporal characteristics that influence active riding choices, such as increased greenery and lower levels of motorization. These insights can inform the planning of street networks and the development of infrastructure to improve the use of active transportation.
\end{abstract}


\end{frontmatter}

\section{Introduction}
\label{sc:intro}

The impact of the built environment on human behavior has long been an area of concern for urban planners, with a particular interest in the transportation planning subfield \citep{eldeeb2021built,yang2018active,lee2016transportation,comendador2014urban,zegras2005sustainable}. The literature points to a growing number of studies that have conducted assessments of environmental factors that affect active transportation \citep{gao2023data,stappersEffectInfrastructuralChanges2018,wangRelationshipNeighborhoodBuilt2017,knuiman2014longitudinal,winters2010built}. Researchers mainly sought to analyze differences in the spatial distribution of bicycle use and to analyze what interventions could be implemented in the built environment to encourage cycling \citep{yangCyclingfriendlyCityUpdated2019}. These efforts have explored many factors influencing people's route choices; for example, researchers have found the quality of cycling infrastructure, perceived safety, accessibility to destinations, and overall cycling experience to play a role in this decision-making process \citep{bao2023understanding,codina2022built,yang2019towards,christiansen2016international,pucherWalkingCyclingUnited2011}.

Rapid advances in deep learning and image analysis techniques have greatly improved the processing of sensory data related to the built environment, particularly through Street View Imagery (SVI)~\citep{biljecki_street_2021}. By sensing the elements and scenes captured in SVI, a more accurate and comprehensive audit of the streetscapes and cycling venues can be performed in comparison to previous approaches \citep{liStreetViewImagery2022}. Recent research focuses primarily on using SVI to assess both physical characteristics and subjective perceptions of the environment \citep{liAssessingStreetlevelUrban2015, ito_understanding_2024, wang2024assessing}. SVI has been applied widely --- in health and urban studies, transportation safety, behavioral preferences, and traffic patterns  --- affirming its tremendous value \citep{biljecki_street_2021,ibrahim2021urban,verhoeven2018differences,li2022underserved,ki2021analyzing,gong2019classifying}. While previous studies have examined associations between real travel data and elements inferred from SVI, trip purposes are often overlooked \citep{li2018investigating,liStreetViewImagery2022,xia2022dutraffic,zhu2024understanding}. However, it has been suggested that incorporating the purpose of the trip into the analysis of the transportation route would inform transportation planning more holistically and effectively \citep{wolf2001elimination,yang2012walking,montini2014trip}, so it is an imperative to give it due consideration.

This study addresses the critical gap in the literature for the first time by pioneering the evaluation of bicycle travel route choices for trip purposes, using SVI and individual bicycle trip data. Our novel approach marks a significant advancement in understanding urban cycling patterns and introduces visual perceptions as variables. In addition, we emphasize the active influence of travel purposes, such as leisure, commuting, and shopping, on cycling behavior. This comprehensive analysis assesses the impact of these factors on road preferences, destination choices, and commute durations, providing a multifaceted understanding of urban cycling dynamics.

\section{Related work}
\label{sc:related_work}
\subsection{Active transportation}

Research on active transportation has gained significant attention in recent years due to its importance in reducing CO$_2$ emissions \citep{brand_climate_2021,2023_epb_semantic_networks}, improving health \citep{mizdrak_assessing_2023}, and bringing economic benefits \citep{gravett_assessing_2021}. Numerous studies have examined various factors that influence active travel behaviors \citep{gotschiComprehensiveConceptualFramework2017}. Among them, the built environment's impact on cycling and walking is a predominant theme, with researchers developing indexes to evaluate bikeability from multiple dimensions. \cite{su16062545} identified five key factors for the design of bicycle infrastructure: safety, comfort, attractiveness, directness, and coherence. Similarly, \cite{ito2021assessing} classified indicators into connectivity, environment, infrastructure, vehicle-cyclist interaction, and perception, while \cite{hartantoDevelopingBikeabilityIndex2017} included transportation status, connectivity, infrastructure, environment, topography, and bicycle parking in their bikeability index.

In terms of specific indicators, land use diversity is a critical factor in active transportation studies. A diverse mix of land uses is believed to reduce travel distances and promote cycling and walking by making destinations more accessible. For example, \cite{zhaoImpactLandUse2020} analyzed the characteristics of land use that affect cycling, finding a positive correlation between the mix of land use and the frequency of cycling. Similarly, \cite{wintersBuiltEnvironmentInfluences2010} found that varied land use compositions correlated with increased cycling trips. Studies have also indicated that mixed land use configurations can encourage prolonged activity levels and higher urban vitality \citep{jacobs-crisioniEvaluatingImpactLandUse2014,guoBuiltEnvironmentEffects2020,noordzijLandUseMix2021}.
Points of interest (POI) also provide detailed information about the destinations and activities available in an area. \citet{carvalhinho_assessment_2024}, for example, evaluated cycling routes in a city in Portugal and reported the importance of points of interest for leisure cycling. POIs and cycling frequency and safety have also been found to have strong associations as well \citep{lee_public_2021, li_road_2024, wu_measuring_2019}. Besides the built environment characteristics, socioeconomic and topographical characteristics have also been found to be associated with cycling \citep{hahn_effect_2001, saglio_examining_2023, vidaltortosa_cycling_2021a, stalling_associations_2022}. 

In terms of methodological approaches, studies on active travel often utilize aggregation and decomposition models. Aggregation models use attributes of the socioeconomic and built environment to form spatial indicators, evaluating the feasibility of active travel on a larger scale \citep{cerinExplainingSocioeconomicStatus2009,wintersBikeScoreAssociations2016,hartantoDevelopingBikeabilityIndex2017,ito2021assessing}. On the other hand, decomposition models explore the relationship between individual travel behaviors and influencing factors using statistical methods such as logistic regression \citep{buehler2012cycling,jaber2021analysis,iamtrakul2023impact,wu2024affects}. For instance, \cite{wuerzer2015cycling} used a questionnaire survey and a random regression model to identify factors that improve bicycle travel competitiveness. Moreover, there has been an increasing number of articles on examining the route choices of active mobility users as it is an important topic to understand what the users prefer and potentially find insights into how to encourage more people to use active transportation \citep{guo_pedestrian_2013, shatu_development_2018b, fitch_road_2020, sarjala_built_2019, brand_impact_2019, cubells_gendered_2023, sevtsuk_role_2022, cubells_escooter_2023, passmore_using_2024}

The analysis of movement trajectories --- often obtained through GPS measurements --- is a recent development that provides nuanced insights into cycling behavior, collecting detailed spatial information including speed, time, and route choice, allowing researchers to visualize and understand movement patterns and interactions with the built environment \citep{marra2019developing,guoBuiltEnvironmentEffects2020,furlettiInferringHumanActivities2013, nelson_crowdsourced_2021}.
Studies have used these data to explore urban traffic flow, analyze route preferences, and infer activities and trip purposes based on trajectory and POI data \citep{o2005geo,titzeDevelopingBikeabilityIndex2012,sunExaminingAssociationsEnvironmental2017,verhoevenDifferencesPhysicalEnvironmental2018,bao2023understanding, lu_understanding_2018, scott_route_2021, hsueh_influential_2023, chung_understanding_2024, ross-perez_identifying_2022, xing_exploring_2020, li_inferring_2021}. However, there is still a scarcity of studies that used GPS trajectories with information on manually labeled trip purposes to investigate the influence of trip purposes on cyclists' trajectories.  

\subsection{Street view imagery}

Street view imagery (SVI) has emerged as a powerful source of data for analyzing the built environment, offering a human eye-level perspective on visual experiences and their impact on travel behaviors. SVI allows for the extraction and reconstruction of visual features in the environment, such as vegetation, sky, and buildings, using advanced deep learning techniques \citep{2023_jag_svi_sensitivity,nagata2020objective,wang2021automatic,xia2021development}. This capability has been widely used for auditing the built environment, assessing street space quality, urban vitality, and temporal changes in urban spaces \citep{alhasounStreetifyUsingStreet2019,wangAssessingStreetSpace2022,liExaminingSpatialDistribution2020,gongClassifyingStreetSpaces2019,maDeepExplorationStreet2023}.

In transportation, SVI has been popular for analyzing walkability and pedestrian characteristics \citep{kim_streetscape_2022, yang_walk_2021a, yang_examining_2023}. For example, \citet{liuStreetViewEnvironments2023} identified factors such as street greenery, land use diversity, and proximity to train stations as contributors to increased walking time on weekends. Additionally, SVI has been used to assess travel safety, focusing on crash frequency and identifying spatial characteristics that influence safety \citep{huInvestigationClustersInjuries2020,stilesHowDoesStreet2022}. Researchers have also explored the use of SVI in analyzing cycling routes, examining factors such as physical environmental characteristics and seasonal variations in cycling behaviors \citep{goelEstimatingCitylevelTravel2018,verhoevenDifferencesPhysicalEnvironmental2018,ritaUsingDeepLearning2023,zhaoEstimatingImpactsSeasonal2024, jeon_effects_2024}. It has also been used to construct bikeability and walkability indexes \citep{koo_how_2022, ito2021assessing}.

Combining GPS trajectories and street view imagery for analysis presents an opportunity for a detailed and multifaceted examination of travel behavior by considering the visual experience of cyclists with high spatial granularity. This integration allows researchers to better understand route choices and environmental interactions in greater depth, addressing research gaps related to travel purposes and providing a robust framework for transportation planning.

\subsection{Research gaps and opportunities}

Despite significant advancements in understanding active transportation and the built environment, several research gaps remain. Previous studies have primarily focused on environmental factors influencing active travel without adequately considering the purposes of these trips. Additionally, there is a lack of detailed comparative spatial studies on the differences between actual travel paths versus the shortest paths in terms of both visual and non-visual factors. Addressing these research gaps is essential for developing a more comprehensive understanding of active transportation behaviors and for informing evidence-based strategies to promote sustainable urban mobility.

This study, therefore, aims to bridge these gaps by analyzing the relationship between bicycle travel purposes, route choices, visual factors (i.e., indicators from SVI), and non-visual factors (e.g., socioeconomic and built environment factors), using GPS trajectory data and SVI. By incorporating trip purposes into the analysis, this research provides a more detailed and nuanced understanding of travel behavior, informing transportation planning and policy-making to promote active transportation effectively.

\section{Data and methods}
\label{sc:data_methods}
\subsection{Overview}
\begin{figure}[H]
\centering
\includegraphics[width=1\linewidth]{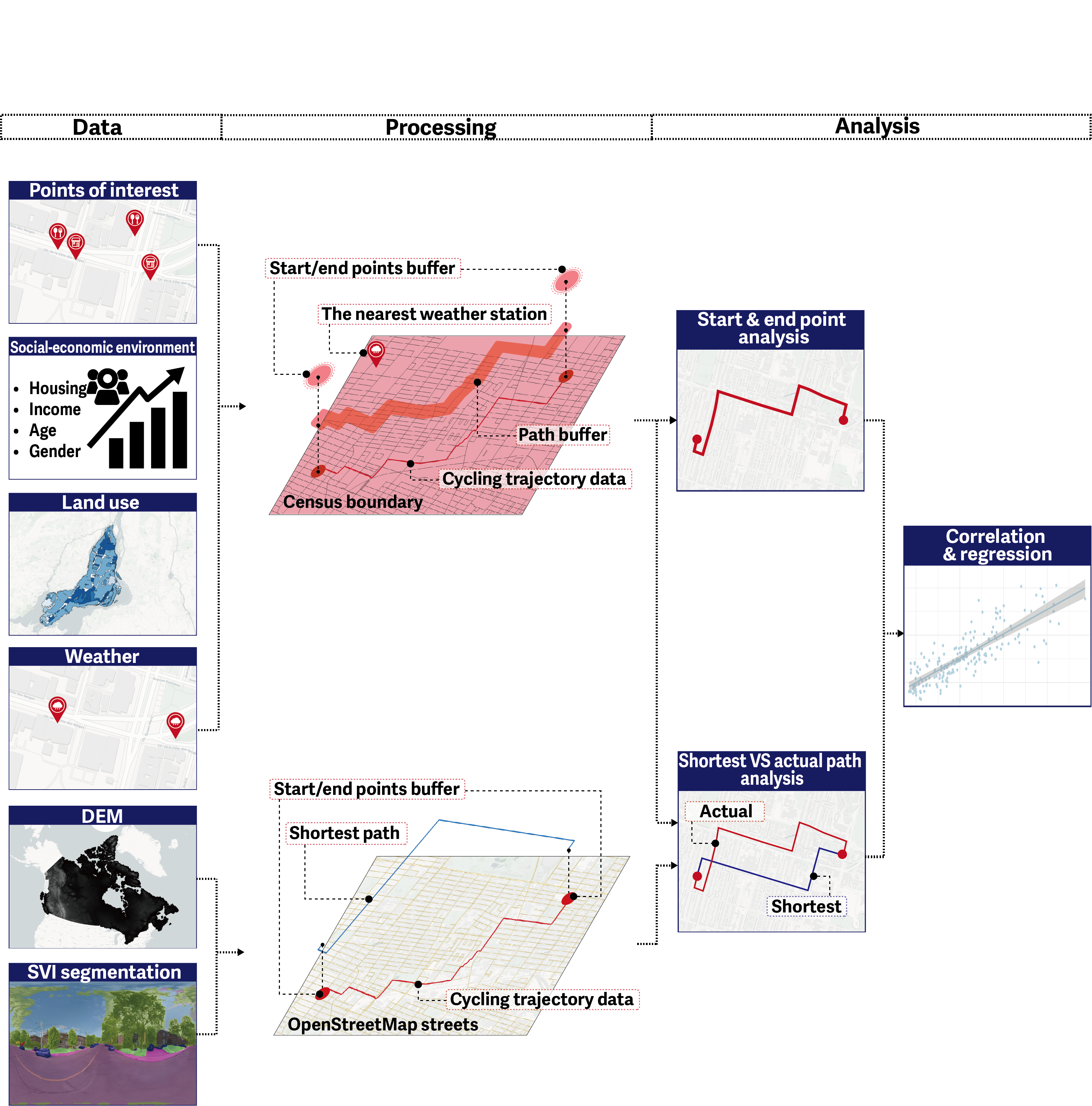}
\caption{Our framework with four phases: 1) data collection, 2) spatial processing, 3) origin-destination and shortest path analysis, and 4) correlation and regression analysis.}
\label{fig:framework}
\end{figure}

We propose a four-stage framework to analyze the relationship between bicycle trip routes, purposes, and environmental characteristics (Figure \ref{fig:framework}). Stage one involves collecting urban bicycle data, including GPS tracks and trip purposes, and analyzing temporal distribution and trip categories. We also gather data on POIs, socioeconomics, land use, weather, topography, and SVI indicators.
Stage two focuses on spatial processing to create a unified dataset. We integrate heterogeneous spatial data into a tabular format and compute shortest paths for all trajectories, joining indicators to both actual and shortest paths. In stage three, we analyze spatial and temporal differences in trip destinations across purposes and examine factors influencing cyclists' route choices.
The final stage involves correlation analyses and regression models to explore interactions between visual and non-visual factors affecting cycling behavior. This framework enables a comprehensive study of urban cycling behaviors.

\subsection{Study area}
Cities are transforming into bicycle-friendly environments to improve livability and promote sustainability. We chose Montreal for its open datasets and bike-friendliness efforts, ranking 18th in the Copenhagen Index \citep{comprehagenize}.
Montreal's cycling initiatives provide an ideal context for studying urban bicycle network construction. The city's 1995 cycling plan integrated existing lanes, repurposed corridors, and public transportation, achieving significant results \citep{cyclingplan}.
The government has implemented various strategies to increase bicycle usage, including the BIXI bike-sharing program \citep{bixi} and cultural events such as the Go Bike Festival \citep{gobikefestival}. According to the Cycling in Quebec 2020 report, cycling in Montreal continues to grow, with 5.0\% of trips to the city center made by bicycle, and some neighborhoods reaching 13.0\% \citep{cycling2020}. This case study offers insights into cyclists' route preferences in a well-established, bike-friendly city.

\subsection{Data}
\subsubsection{Cycling trajectory data}
This study uses data from OpenStreetMap, street view imagery, and government sources. Individual trips are recorded by Montreal's MyVeVelo application, providing user IDs, times, paths, lengths, and destinations from 2013 to 2015. The city processed the data to address GPS accuracy issues before public release \citep{dataresource}.
The analysis focuses on cyclist-chosen roads rather than all city roads. While sample representativeness cannot be confirmed, our primary aim is to examine relationships between trip purpose, spatial characteristics, and perceived bicycle appropriateness in active travel, minimizing the impact of this limitation.
We selected factors to assess trip differences, categorized into visual and non-visual factors \citep{desjardins2021going,mcmains2011interactions,juarez2023cyclists}. Visual factors relate to perceivable changes in the street environment during cycling, while non-visual factors include neighborhood-level indicators such as socioeconomic status and built environment characteristics affecting cyclists' behavior.

\subsubsection{Variable construction}
\paragraph{Non-visual factors} Based on previous research, we selected 14 control variables related to the natural and built environment based on the availability and accessibility of data to form the non-visual database for this study (see Table \ref{tab:control_variables}). One of the variables is the average slope calculated from DEM elevation data within each travel trajectory. We also used socioeconomic control variables such as housing price, after-tax income, gender differences, and average age, which were aggregated with different levels of census boundary results and spatially joined with bicycle trajectory data. Specifically, gender differences were made using the proportion of males in the overall sex. Such publicly available data hosted by the City of Montreal come from the 2016 Census results. The average temperature of the day was another variable, which was calculated for each travel trajectory based on the meteorological data obtained from the nearest meteorological station. Based on the trip time recorded in the original trip data, we also obtained the month, day of the week, and start hour of the trip based on the start time. At the same time, the start time, end time, and path length were used to generate the overall travel speed. To simplify the analysis, the day of the trip was categorized as weekdays and weekends, assigning a value of 0 to weekdays and 1 to weekends. 

\begin{table}[H]
    \centering
    \caption{Description of variables used in this study.}
    \resizebox{\textwidth}{!}{%
    \begin{tabular}{>{\raggedright\arraybackslash}p{4cm} >{\raggedright\arraybackslash}p{3cm} >{\raggedright\arraybackslash}p{4cm} >{\raggedright\arraybackslash}p{6cm}}
        \toprule
        Type & Category & Data Source & Description \\
        \midrule
        \multirow{8}{*}{\textbf{Non-visual variables}} & Land use & open.canada.ca & land use mix \\
        & Slope & open.canada.ca & mean value of trajectory slope \\
        & Temperature & open.canada.ca & temperature of the day (degrees Celsius) \\
        & Gender difference & statcan.gc.ca & male proportion (dissemination areas) \\
        & Average age & statcan.gc.ca & average (dissemination areas) \\
        & POIs & openstreetmap & count (radius 100m) \\
        & Housing price & statcan.gc.ca & average (census subdivisions) \\
        & After-tax income & statcan.gc.ca & mean (census tracts) \\
        \midrule
        \multirow{8}{*}{\textbf{Visual variables}} & Road & street view imagery & pixel ratios of road \\
        & Sidewalk & street view imagery & pixel ratios of sidewalks \\
        & Building & street view imagery & pixel ratios of buildings \\
        & Greenery & street view imagery & pixel ratios of vegetation \\
        & Sky & street view imagery & pixel ratios of sky \\
        & Motorization & street view imagery & pixel ratios of cars, trucks, buses, trains, motorcycles \\
        & Active mobility & street view imagery & pixel ratios of people, riders, bicycles \\
        & Obstruction & street view imagery & pixel ratios of walls, buildings, poles \\
        \midrule
        \multirow{7}{*}{\textbf{Cyclist trajectory data}} & Day of week & the City of Montreal & weekdays and weekends \\
        & Hour & the City of Montreal & counting by track start time \\
        & Month & the City of Montreal & counting by track start month \\
        & Duration & the City of Montreal & difference between the start and end times of the trajectory (mins) \\
        & Length & the City of Montreal & the length of the trajectory (meter) \\
        & Speed & the City of Montreal & average speed of the trip (m/s) \\
        \bottomrule
    \end{tabular}%
    }
    \label{tab:control_variables}
\end{table}

As the impact of land use on active travel behavior has been shown in \autoref{sc:related_work}, we included the land use mix to represent the degree of diversity among land use within a 500-meter radius from the origins and destinations. To estimate the land use mix, we selected the Shannon Diversity Index as the estimation method \citep{nolan2006beachcomber}, which is often used to estimate the level of diversity of a group and is mostly used in species richness studies, which is calculated as:
\begin{equation} \label{eq:shannon}
Shannon = - \sum_{i=1}^{n} p_i \log(p_i)
\end{equation}
where $p_i$ is the proportion of individuals of species $i$. 

Furthermore, after the exploratory analysis of the trip purpose for the raw data, the trip purposes were summarized into eight main categories: commuting, work-related, social, shopping, school, sports, leisure, and other. In the literature, it has been reported that the purpose of cycling is closely correlated to the POIs of the place \citep{furlettiInferringHumanActivities2013}. Thus, we downloaded the following facilities and amenities from OpenStreetMap (OSM) as POIs: offices, recreational facilities, cycling facilities, restaurants, and tourist points of interest, which were then used to calculate the total number of POIs within a radius of 100 meters from the destinations.
In this research we relied on OSM to acquire different types of features. 
While OSM tends to have variable levels of quality, POIs and cycling infrastructure are mapped well~\citep{ferster2020,Viero2024,McCarty_2023,Pinho_2023,Zhang_2019}, especially in Quebec~\citep{Moradi2021}, and have been used widely in research. 

\paragraph{Visual factors} 
To analyze visual factors, we extracted semantic information from street view imagery (SVI) corresponding to GPS points along each travel path. Our methodology involved inputting these trajectory points into the Google Street View (GSV) Application Programming Interface (API) to obtain associated panoramic images, each with a unique identifier, latitude, longitude, and capture period (year and month). To ensure data accuracy and relevance, we acquired street view images based on sampling points closest to the recorded GPS locations. Furthermore, we constrained the temporal range of the imagery to span from 2013 to 2015, aligning with our study period.

Subsequently, we applied semantic segmentation on the collected SVI with the state-of-the-art model called Mask2Former pre-trained on Cityscapes with the mean intersection over union (mIoU) of 84.5\% among 19 classes \citep{cordts2016cityscapes, cheng_maskedattention_2022} by using a Python package called ZenSVI \citep{ito_zensvi_2024}. We strategically selected five individual semantic classes' pixel ratios as indicators: road, sidewalk, building, greenery, and sky. These classes were chosen based on their significant influence on urban perception and mobility, as established in previous urban studies. These classes were chosen based on their significant influence on urban perception and mobility, as established in previous urban studies \citep{biljecki_street_2021}. To capture more complex urban characteristics, we also created three multi-class indices by aggregating multiple semantic classes: motorization, active mobility, and obstruction (see Table \ref{tab:control_variables}). The pixel ratio for each class and index was calculated by computing the proportion of pixels for the respective semantic class or classes after segmentation.

Motorization refers to the interference of vehicles with bicycling, which is calculated using five semantic segmentation metrics:
\begin{equation} \label{eq:motor}
{\text{the degree of motorization}} =  R_\text{car} + R_\text{trucks} +R_\text{buses} +R_\text{trains} + R_\text{motorcycle}
\end{equation}
where $R$ represents the proportion of different semantic segmentation types in image recognition.

Active mobility refers to the presence of pedestrians and cyclists on the street. This indicator can reflect the characteristics and conditions of the space, including but not limited to the level of activity of urban life, the characteristics of social activity on the street, and the characteristics of street traffic:
\begin{equation} \label{eq:active}
{\text{the degree of active mobility}} = R_\text{people} + R_\text{rider} + R_\text{bicycles}
\end{equation}

The degree of obstruction refers to the compactness of the space and the smoothness of passage, and can reflect the degree of restriction caused by buildings, walls, and poles on both sides of the street for cycling:
\begin{equation} \label{eq:obstruct}
{\text{the degree of obstruction}} =  R_\text{walls} + R_\text{buildings} +R_\text{poles}
\end{equation}

The selection of these specific classes and the creation of multiclass indices were motivated by their potential to comprehensively represent the urban environment's key features. For instance, the `motorization' index combines elements related to vehicular infrastructure, while `active mobility' encompasses features conducive to walking and cycling. The `obstruction' index aggregates elements that might impede movement or visibility in the urban landscape. This approach allows us to not only analyze individual urban elements but also to capture the street's complex mix of features that can influence users' mental perception of the environment.

\subsection{Methods}
We structured our analysis in two parts --- 1) origin and destinations and 2) route choice analysis --- to understand the relationships between the trip purposes, the starting and ending points, and the route preferences by cyclists. Each part was then conducted in two steps --- 1) bivariate analysis and 2) multivariate regression analysis --- to examine the data comprehensively.

\subsubsection{Origin and destination analysis}
Our analysis in this part focuses on the impact of non-visual spatial factors on overall bicycle travel patterns. We aimed to access data at the finest available granularity to capture nuanced spatial variations. Importantly, we sought to investigate potential disparities in social factors between trip origins and destinations by following the approach by \citep{furlettiInferringHumanActivities2013}. We denoted those variables collected around the origins with $.x$ suffix and those collected around the destinations with $.y$ suffix. We listed the variables used in this analysis in Table \ref{tab:observe}. 

\begin{table}[H]
    \centering
    \caption{Descriptive statistics of starting and ending points for non-visual analysis.}
    \resizebox{\textwidth}{!}{%
    \begin{tabular}{>{\raggedright\arraybackslash}p{4cm} >{\raggedright\arraybackslash}p{5cm} >{\raggedright\arraybackslash}p{5cm} >{\raggedright\arraybackslash}p{6cm}}
        \toprule
        Category & Name & Location & Comments \\
        \midrule
        \multirow{2}{*}{Land use} & land use.x & start point of trip & \multirow{2}{*}{500m buffer for aggregation} \\  
         & land use.y & endpoint of trip &  \\\midrule
        \multirow{2}{*}{Housing price} & housing price.x & start point of trip & \multirow{2}{*}{the census subdivisions} \\ 
        & housing price.y & endpoint of trip &  \\\midrule
        \multirow{2}{*}{After tax income} & after-tax income.x & start point of trip & \multirow{2}{*}{the census tracts} \\ 
        & after-tax income.y & endpoint of trip &  \\\midrule
        \multirow{2}{*}{Temperature} & temperature.x & start point of trip & \multirow{2}{*}{the nearest meteorological station} \\ 
        & temperature.y & endpoint of trip &  \\\midrule
        \multirow{2}{*}{Sex difference} & male proportion.x & start point of trip & \multirow{2}{*}{the dissemination areas} \\ 
        & male proportion.y & endpoint of trip &  \\\midrule
        \multirow{2}{*}{Average age} & average age.x & start point of trip & \multirow{2}{*}{the dissemination areas} \\ 
        & average age.y & endpoint of trip &  \\
        \bottomrule
    \end{tabular}%
    }
    \label{tab:observe}
\end{table}

In the bivariate analysis, to understand the relationship between characteristics of the environment around the origins and destinations, we first conducted a correlational analysis by computing Pearson's r for each combination of variables. Subsequently, we conducted a descriptive analysis to investigate the distributions of different variables among different trip purposes. In the multivariate regression analysis, we used the analysis of variance (ANOVA) test to check whether the variance of non-visual variables is statistically different among different trip purposes. The model was defined as follows:

\begin{equation} \label{eq:anova_1}
\begin{aligned}
\log\left(\frac{P(Y_i = j)}{P(Y_i = J)}\right) = \beta_{0j} &+ \underbrace{\sum_{k=1}^{4} \beta_{kj} X_{ki}}_{\text{Travel trajectory-related}} + \underbrace{\sum_{m=1}^{11} \gamma_{mj} Z_{mi}}_{\text{Non-visual}} \\
\end{aligned}
\end{equation}

Where:
$J$ is the reference category (Work-related), and $j$ represents each category of the outcome variable (Commute, School, Sport, Leisure, Shopping, Social, Others). Travel trajectory-related variables are:
\begin{itemize}
  \item $X_{1i} = \text{length}_i$
  \item $X_{2i} = \text{speed}_i$
  \item $X_{3i} = \text{du}_i$
  \item $X_{4i} = \text{compare\_path}_i$
\end{itemize}

Recent scholarly analyses have focused on identifying the characteristics of the cycling network that cause cyclists to deviate from the shortest path \citep{lu2018understanding,vorster2018cycle,krizek2007detailed,rupi2018evaluating}. Building on this foundation, our study introduces a quantitative measure to assess this deviation. In the first part of our overall data analysis, we examine the discrepancy between the shortest path (derived from OpenStreetMap) and the actual path taken by cyclists. To quantify this deviation and measure route efficiency, we defined and included a variable called ``compare path," calculated as follows:
\begin{equation} \label{eq:compare}
\text{compare path} = \frac{length_{\text{actual path}} - length_{\text{shortest path}}}{length_{\text{shortest path}}}
\end{equation}

Non-visual variables include:
\begin{itemize}
  \item $Z_{1i} = \text{slope}_i$
  \item $Z_{2i} = \text{temperature}_i$
  \item $Z_{3i} = \text{male\_proportion}_i$
  \item $Z_{4i} = \text{after\_tax\_income}_i$
  \item $Z_{5i} = \text{hour}_i$
  \item $Z_{6i} = \text{POIs}_i$
  \item $Z_{7i} = \text{average\_age}_i$
  \item $Z_{8i} = \text{housing\_price}_i$
  \item $Z_{9i} = \text{month}_i$
  \item $Z_{10i} = \text{day\_of\_week}_i$
  \item $Z_{11i} = \text{lum}_i$
\end{itemize}

\subsubsection{Route choice analysis}
The route choice analysis examines the influence of various trip purposes on cycling path selection, incorporating both socioeconomic factors and streetscape environmental differences at the street level. This analysis focuses on the non-overlapping portions between the shortest paths and the actual paths taken by cyclists. A radial range difference of 20 meters is used to define the non-overlapping segments. For the non-overlapping path segments, socioeconomic elements are averaged based on the information from all spaces through which the path passes. These elements include the spatial distribution of income from census tracks, housing prices from census subdivisions, and age and gender differences from dissemination areas. To account for the consistent spatial orientation of the non-overlapping path, the spatial extent for data acquisition was adjusted. While a 500-meter radius was used for start and end points, a narrower 50-meter radial along the path was employed to obtain a more accurate degree of land use mix.

The analysis of street view images along the paths involves a two-step process. First, visual difference characteristics for all spatial points in the study area are obtained by clustering the points of all paths. Second, the street view static API is used to acquire panoramic images of the nearest neighbor for each point obtained from the trajectory. In the data aggregation process for each path, a 10-meter radial range is used to acquire points. The different metrics derived from street view segmentation are then averaged for these points, providing a comprehensive view of the streetscape along the chosen routes.

In the bivariate analysis, we investigated the differences in the distributions of visual and non-visual variables between actual and shortest paths for different purposes to examine their potential associations. In the multivariate regression analysis, we used the ANOVA test with travel trajectory-related variables, visual variables, and non-visual variables to analyze the statistical differences in variance among different trip purposes. The model is similar to \autoref{eq:anova_1}

\begin{equation} \label{eq:anova_2}
\begin{aligned}
\log\left(\frac{P(Y_i = j)}{P(Y_i = J)}\right) = \beta_{0j} &+ \underbrace{\sum_{k=1}^{4} \beta_{kj} X_{ki}}_{\text{Travel trajectory-related}} + \underbrace{\sum_{m=1}^{11} \gamma_{mj} Z_{mi}}_{\text{Non-visual}} + \underbrace{\sum_{n=1}^{9} \delta_{nj} V_{ni}}_{\text{Visual}} \\
\end{aligned}
\end{equation}

Where:
all the denotations are the same as \autoref{eq:anova_1} and visual variables ($V_{ni}$) include:
\begin{itemize}
  \item $V_{2i} = \text{road}_i$
  \item $V_{3i} = \text{sidewalk}_i$
  \item $V_{4i} = \text{building}_i$
  \item $V_{5i} = \text{vegetation}_i$
  \item $V_{6i} = \text{sky}_i$
  \item $V_{7i} = \text{active\_mobility}_i$
  \item $V_{8i} = \text{obstruction}_i$
  \item $V_{9i} = \text{motorization}_i$
\end{itemize}

\section{Results}
\label{sc:results}
\subsection{Origin and destination analysis}
\subsubsection{Bivariate analysis}
\begin{figure}[h]
\centering
\includegraphics[width=0.8\linewidth]{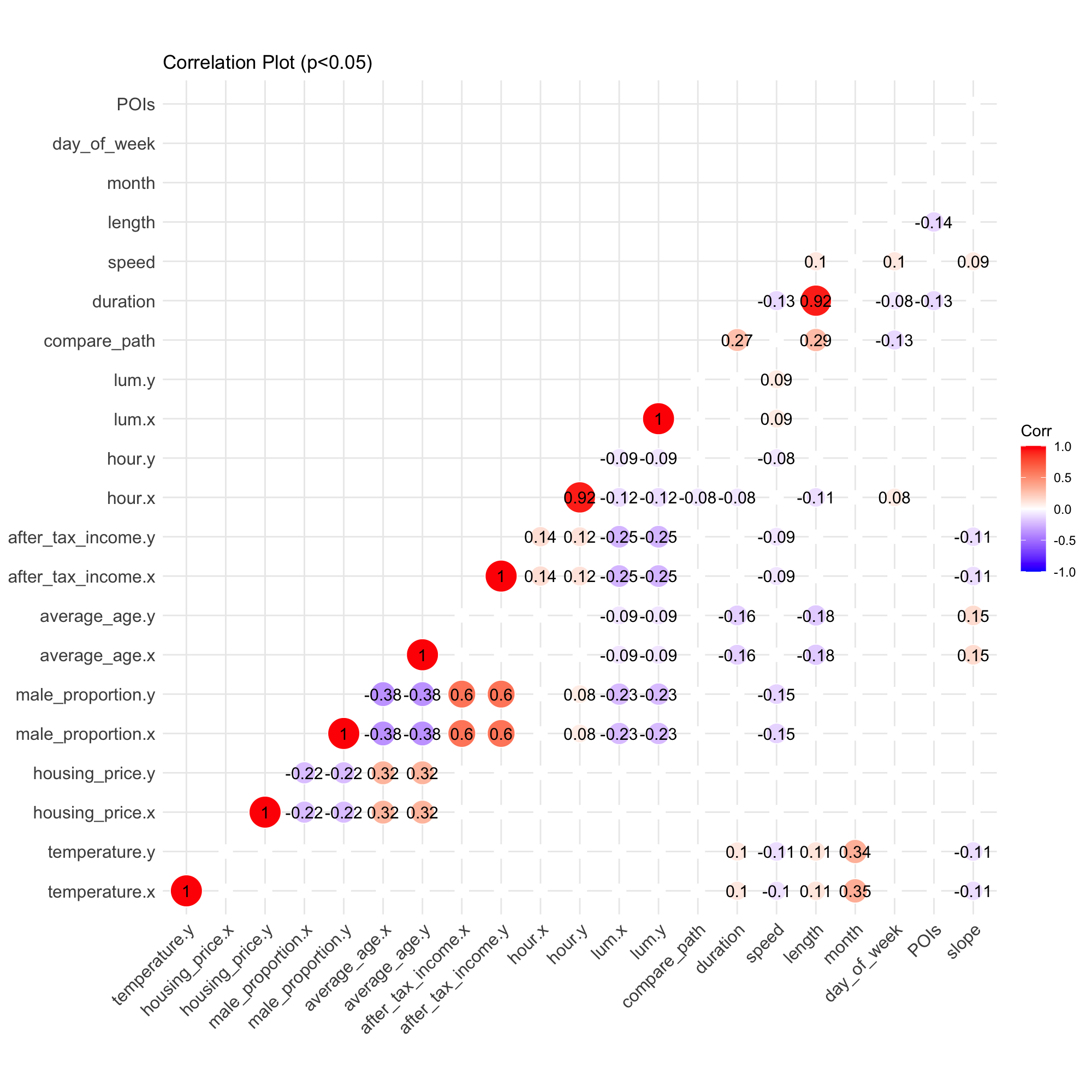}
\caption{Correlation matrix of control variables for non-visual analysis, including the difference between starting and ending points of trips.}
\label{fig:correlation_matrix}
\end{figure}

Correlation analysis of bicycle travel data can provide multiple insights to help understand the relationships between different elements and their impact on travel behavior. \autoref{fig:correlation_matrix} illustrates the correlation matrix for obtaining results for non-visual variables. It is notable that socioeconmic characteristics around the origin and destination points are perfectly correlated --- this could be because many trips were made within the same census boundaries. Based on this observation, we used only the variables around the origins for the subsequent analysis.

In addition, some interesting social phenomena can be found among socioeconomic variables in the origins of cycling trips. For example, the male gender share is positively correlated with after-tax income, with a correlation as high as 0.6, but there is a significant negative correlation between it and the average age and the degree of land use mix, at -0.38 and -0.23, respectively. The median house price in the area where active trips occur is also related in some way to gender variance, with a higher share of men in the area being associated with a lower median house price. Additionally, we also found associations between travel trajectory variables. There is a correlation of 0.29 and 0.27 between compare\_path (i.e., deviation from the shortest path) and trip length and duration, respectively, indicating that the longer the trip is, the more cyclists deviate from the shortest paths. This finding aligns with the findings in the study of \citet{bongiorno_vectorbased_2021} that used GPS trajectories of pedestrians.

\begin{figure}[H]
    \centering
    \begin{subfigure}[t]{0.5\textwidth}
        \centering
        \includegraphics[width=\textwidth]{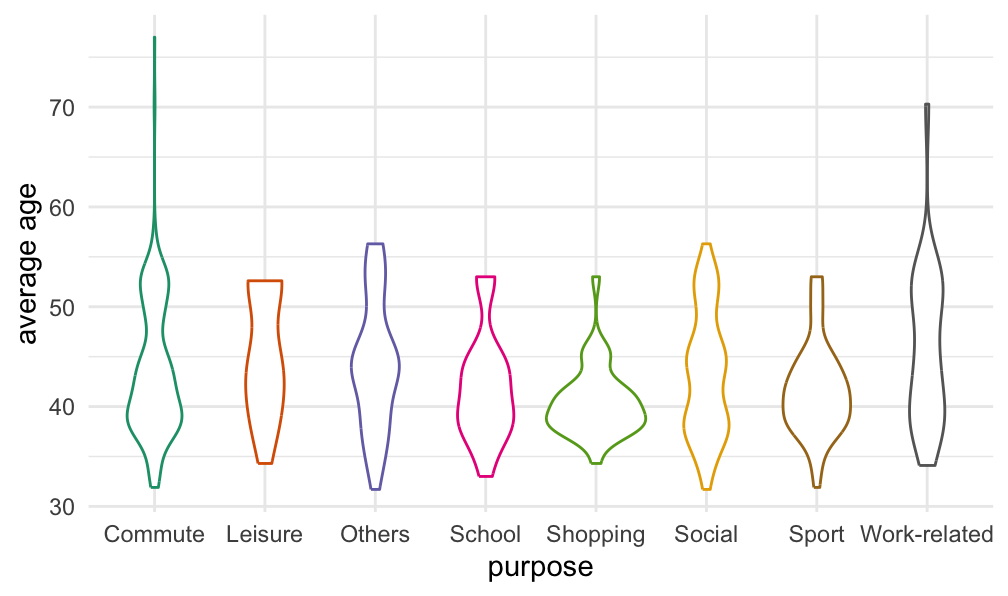}
        \caption{Age distribution by cycling purpose.}
        \label{fig:age}
    \end{subfigure}%
    \begin{subfigure}[t]{0.5\textwidth}
        \centering
        \includegraphics[width=\textwidth]{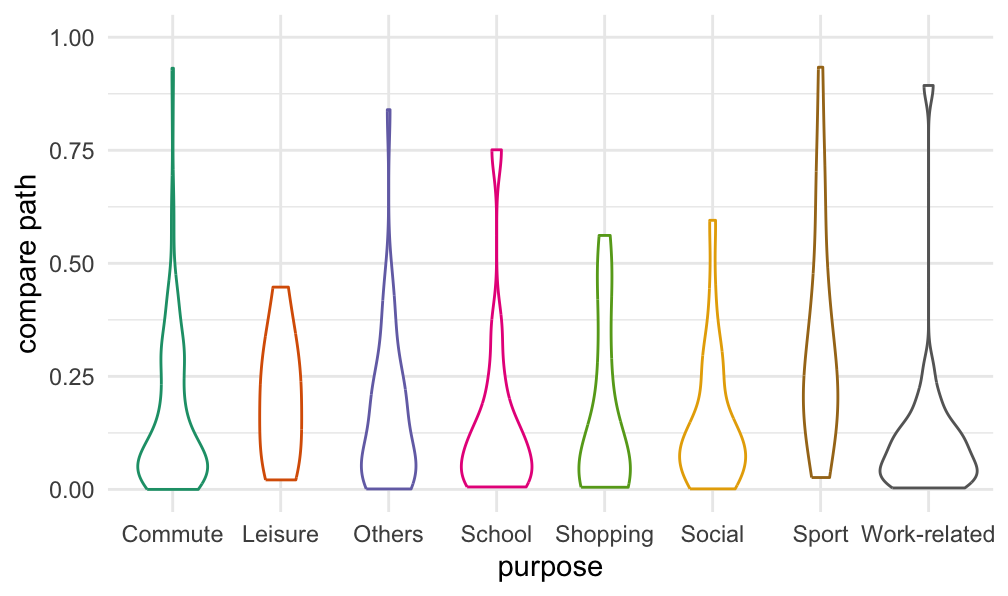}
        \caption{Route choosing variability (\%) distribution by cycling purpose}
        \label{fig:comparep}
    \end{subfigure}

    \begin{subfigure}[t]{0.5\textwidth}
        \centering
        \includegraphics[width=\textwidth]{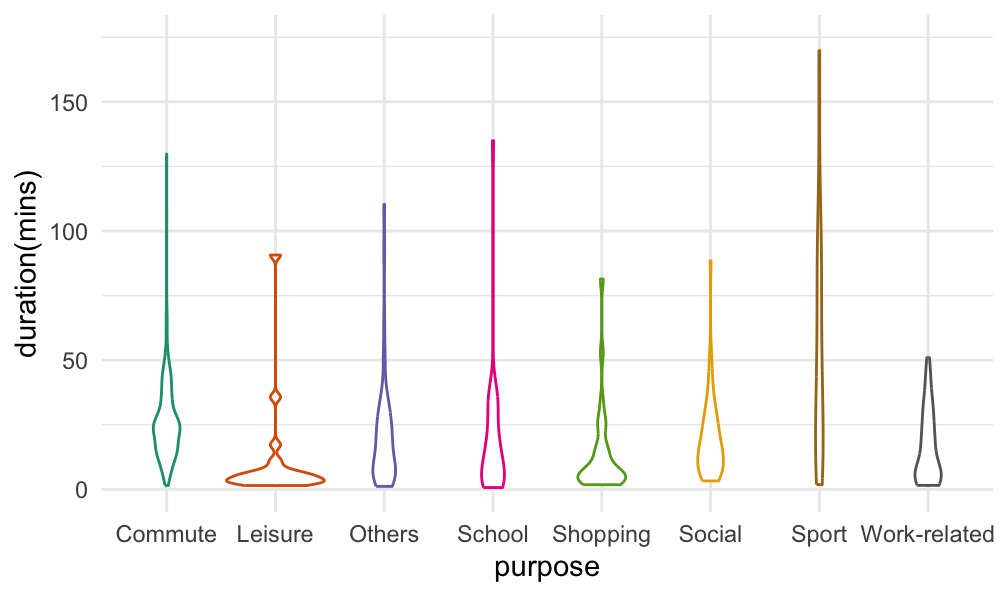}
        \caption{Duration(min) distribution by cycling purpose}
        \label{fig:duration}
    \end{subfigure}%
    \begin{subfigure}[t]{0.5\textwidth}
        \centering
        \includegraphics[width=\textwidth]{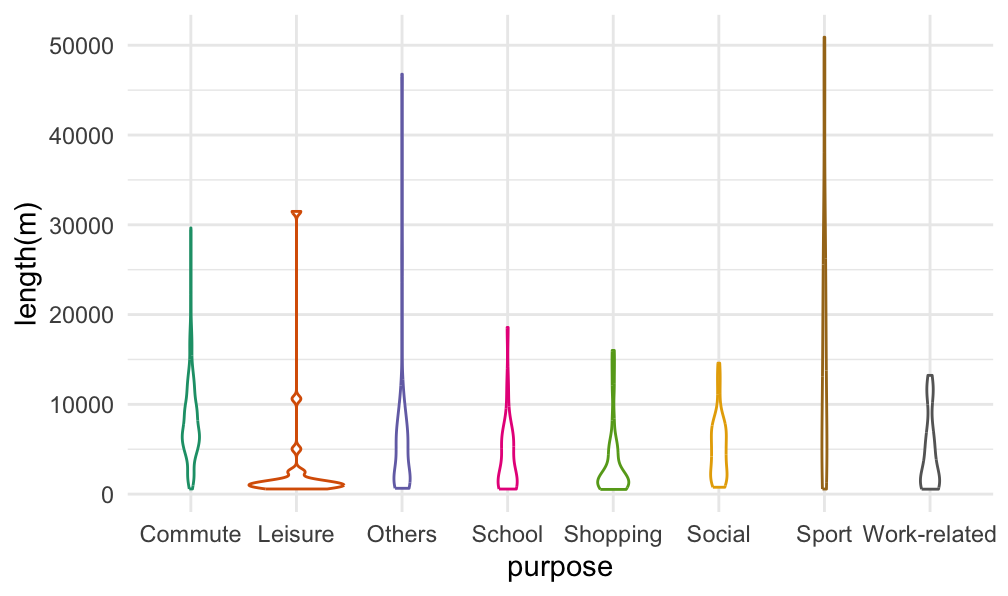}
        \caption{Trip distance(m) distribution by cycling purpose}
        \label{fig:distance}
    \end{subfigure}

    \begin{subfigure}[t]{0.5\textwidth}
        \centering
        \includegraphics[width=\textwidth]{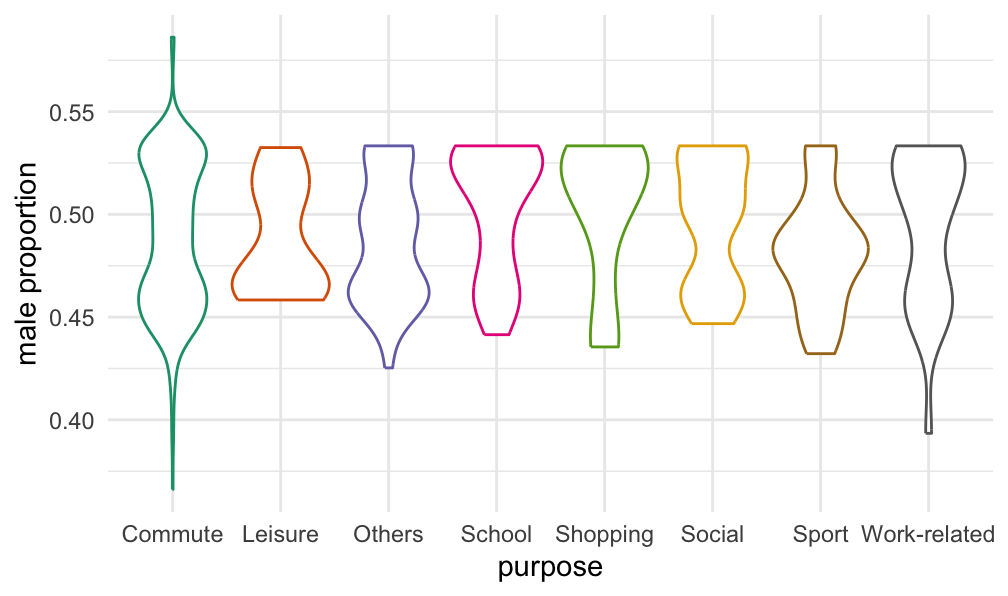}
        \caption{Male proportion distribution by cycling purpose}
        \label{fig:gender}
    \end{subfigure}%
    \begin{subfigure}[t]{0.5\textwidth}
        \centering
        \includegraphics[width=\textwidth]{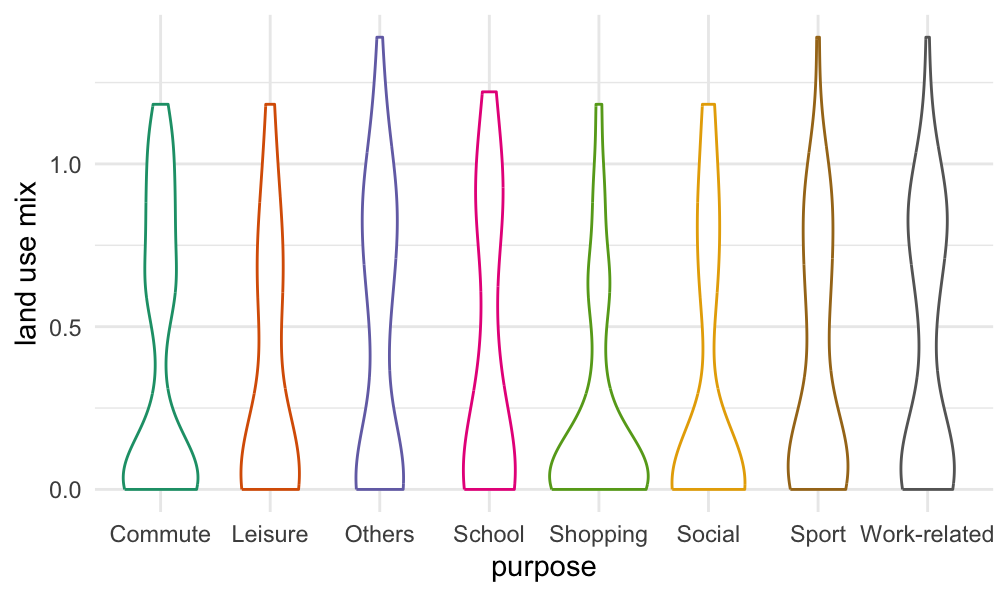}
        \caption{Land use mix distribution by cycling purpose}
        \label{fig:lum}
    \end{subfigure}

    \begin{subfigure}[t]{0.5\textwidth}
        \centering
        \includegraphics[width=\textwidth]{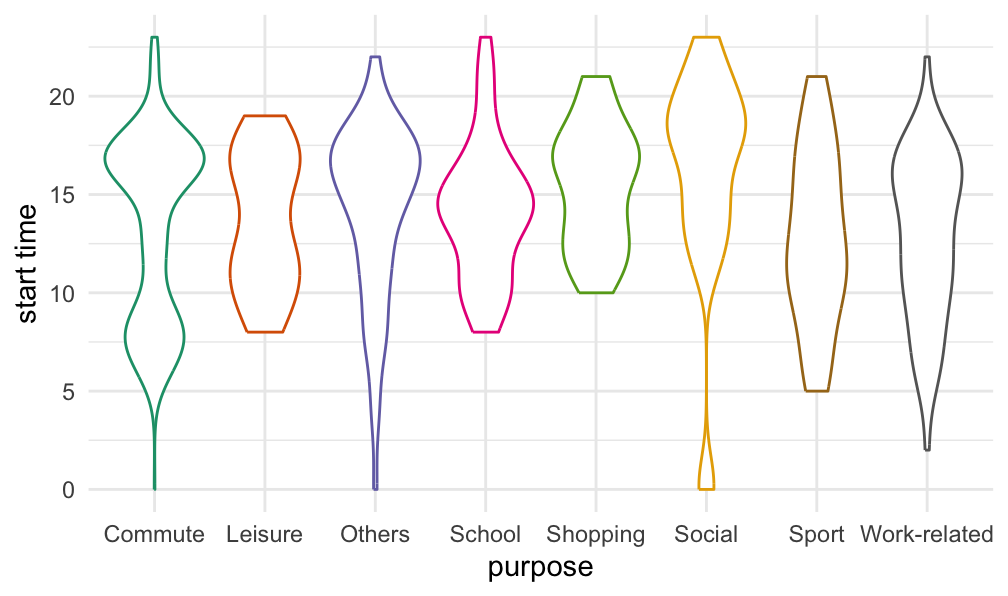}
        \caption{Trip's start time distribution by cycling purpose}
        \label{fig:start_time}
    \end{subfigure}%
    \begin{subfigure}[t]{0.5\textwidth}
        \centering
        \includegraphics[width=\textwidth]{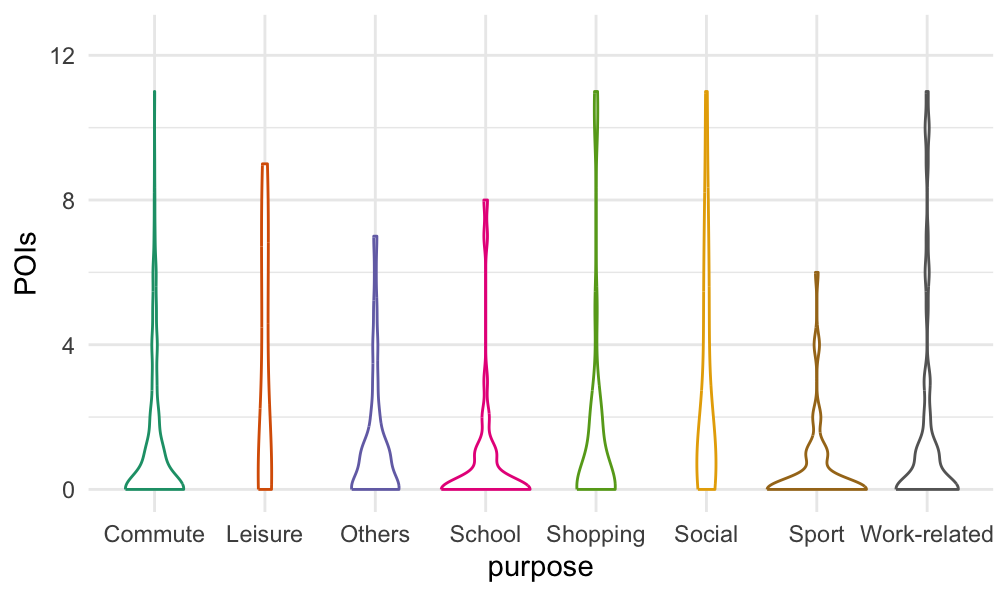}
        \caption{Distribution of POIs density by cycling purpose}
        \label{fig:poi}
    \end{subfigure}
    
    \caption{Distributions of various factors by cycling purpose}
    \label{fig:all_distributions}
\end{figure}

To gain valuable insights into data distribution across different trip purposes, we generated violin plots for non-visual variables. \autoref{fig:age} reveals intriguing patterns in the average age distribution at trip origins. Shopping-related trips exhibit the most concentrated age distribution, centering around 39 years, which also represents the lowest median age for all purposes. In contrast, leisure, social, and work-related purposes tend to attract older demographics, showcasing a wider age range and higher median age. These distinctions in age profiles for trip purposes offer nuanced insights into the demographic factors that influence cycling behaviors for various activities.

The deviation between actual and shortest paths varies significantly across trip purposes, with sports-related journeys exhibiting the most pronounced differences (\autoref{fig:comparep}). For exercise trips, the actual route length often substantially exceeds the shortest path, indicating a preference for longer, potentially more scenic or challenging routes. This pattern contrasts sharply with commute, school, and work-related trips, which demonstrate much smaller deviations from the shortest paths. Sports-related cycling also stands out in terms of duration and distance, consistently surpassing other trip categories (\autoref{fig:duration}, \autoref{fig:distance}). These patterns suggest that sports and exercise cyclists prioritize the journey itself, often choosing extended routes that may offer better training opportunities or more enjoyable experiences, rather than focusing solely on efficient destination-reaching. Interestingly, while leisure trips are also undertaken for enjoyment, they exhibit significantly shorter durations and distances compared to sports-related cycling. This distinction highlights the diverse nature of recreational cycling, where leisure rides prioritize casual enjoyment over physical exertion, potentially catering to a broader range of cyclists with varying fitness levels and time constraints.

\autoref{fig:gender} depicts the distribution of the male proportion for cycling trip origins based on various trip purposes. Notably, commuting and work-related trips exhibit the widest range of male proportions, suggesting greater variability compared to other purposes. Commute, school, and shopping trips display a slightly higher male proportion, hovering around 0.525, whereas social trips have proportions closer to an equal gender split, around 0.50. Conversely, purposes such as others, leisure, and sports show lower male proportions, indicating a tendency towards a higher female representation for these activities.

Moreover, we found a notable difference in the level of land mix at the start of the trips, with work-related trips showing a higher level of land mix compared to other classifications, while leisure and social trip origins have the lowest degree of land mix (\autoref{fig:lum}). This pattern suggests that the diversity of land use at trip origins may influence cycling behavior, with work-related trips potentially benefiting from or being more common in areas with varied urban functions.

The temporal distribution of cycling trips exhibits distinct patterns across different purposes (\autoref{fig:start_time}). Commuting trips are heavily concentrated during traditional rush hours, while work-related journeys predominantly occur throughout the workday. Shopping and social activities demonstrate a notable afternoon peak, in contrast to exercise trips, which reach their zenith in the evening hours. Spatial analysis reveals that leisure and social trip endpoints are associated with significantly higher densities of points of interest (POIs) compared to other trip categories (\autoref{fig:poi}), suggesting a relationship between these activities and areas of diverse urban amenities. Interestingly, several variables including after-tax income, temperature, month of the year, route slope, and cycling speed show no significant variations across trip purposes, indicating that these factors may have a more uniform influence on cycling behavior regardless of the journey's intent.

\subsubsection{Multivariate regression analysis}

\autoref{tab:anova} presents the results of an ANOVA test for multiple regression models, using trip purpose as the target variable. This analysis reveals significant variations in the influence of different controlling factors on trip purposes. Factors demonstrating high statistical significance (p < 0.001) include trip length, hour of the day, POI density, day of the week, and the difference between actual and shortest path lengths. Additionally, several factors show moderate significance (p < 0.01 or p < 0.05), such as trip duration, speed, gender proportion, average age, and land use mix (LUM). Interestingly, some factors previously considered important, such as temperature, slope, after-tax income, housing price, and month of travel, did not show statistically significant relationships with trip purpose in this model (p > 0.05). This aligns with our findings in the bivariate analysis and suggests that while these factors may influence overall cycling behavior, they may not be as crucial in differentiating between trip purposes.

The high significance of temporal factors (hour of the day and day of the week) and spatial factors (POI density and path comparison) underscores the importance of considering both time and space in understanding cycling trip purposes. The significance of demographic factors such as gender proportion and average age also highlights the role of social characteristics in shaping trip purposes.
These findings provide valuable insights for urban planners and policymakers aiming to promote and facilitate different types of bicycle trips. By focusing on the most influential factors, targeted strategies can be developed to enhance cycling infrastructure and encourage diverse cycling activities across various trip purposes.

\begin{table}[H]
    \centering
    \caption{Non-visual analysis: ANOVA test for multiple regression models based on trip purpose.}
    \resizebox{\textwidth}{!}{%
    \begin{threeparttable}
    \begin{tabular}{>{\raggedright\arraybackslash}p{4cm} >{\raggedright\arraybackslash}p{2cm} >{\raggedright\arraybackslash}p{3cm} >{\raggedright\arraybackslash}p{3cm} >{\raggedright\arraybackslash}p{3cm} >{\raggedright\arraybackslash}p{3cm}}
        \toprule
        Category & Df & Deviance & Resid. Df & Resid. Dev & Pr(>F) \\
        \midrule
        Length & 7 & 43.21 & 4557 & 1702.4 & 3.04e-07 *** \\
        Temperature & 7 & 8.23 & 4557 & 1667.4 & 0.31 \\
        Duration & 7 & 22.56 & 4557 & 1681.7 & 0.00 ** \\
        Speed & 7 & 16.54 & 4557 & 1675.7 & 0.02 * \\
        Slope & 7 & 6.74 & 4557 & 1665.9 & 0.45 \\
        Male proportion & 7 & 17.53 & 4557 & 1676.7 & 0.01 * \\
        After-tax income & 7 & 9.79 & 4557 & 1669.0 & 0.20 \\
        Hour & 7 & 29.58 & 4557 & 1688.8 & 0.00 *** \\
        POIs & 7 & 32.74 & 4557 & 1691.9 & 2.96e-05 *** \\
        Average age & 7 & 19.91 & 4557 & 1679.1 & 0.01 ** \\
        Housing price & 7 & 6.53 & 4557 & 1665.7 & 0.48 \\
        Month & 7 & 11.39 & 4557 & 1670.6 & 0.12 \\
        Day of week & 7 & 77.23 & 4557 & 1736.4 & 5.04e-14 *** \\
        Lum & 7 & 16.68 & 4557 & 1675.9 & 0.02 * \\
        Compare path & 7 & 34.78 & 4557 & 1694.0 & 1.23e-05 *** \\
        \bottomrule
    \end{tabular}
    \begin{tablenotes}
        \footnotesize
        \item \textit{Significance codes:} 0 ‘***’ 0.001 ‘**’ 0.01 ‘*’ 0.05 ‘.’ 0.1 ‘ ’ 1
    \end{tablenotes}
    \end{threeparttable}
    }
    \label{tab:anova}
\end{table}

\subsection{Route choice analysis}
This section analyzes the differences between actual routes and shortest paths for various trip purposes. We examine how specific control variables influence route choices and how these factors vary across different trip purposes. This analysis reveals important insights into travel behaviors and route selection patterns.

\subsubsection{Bivariate analysis}
\begin{figure}[H]
\centering
\includegraphics[width=1\linewidth]{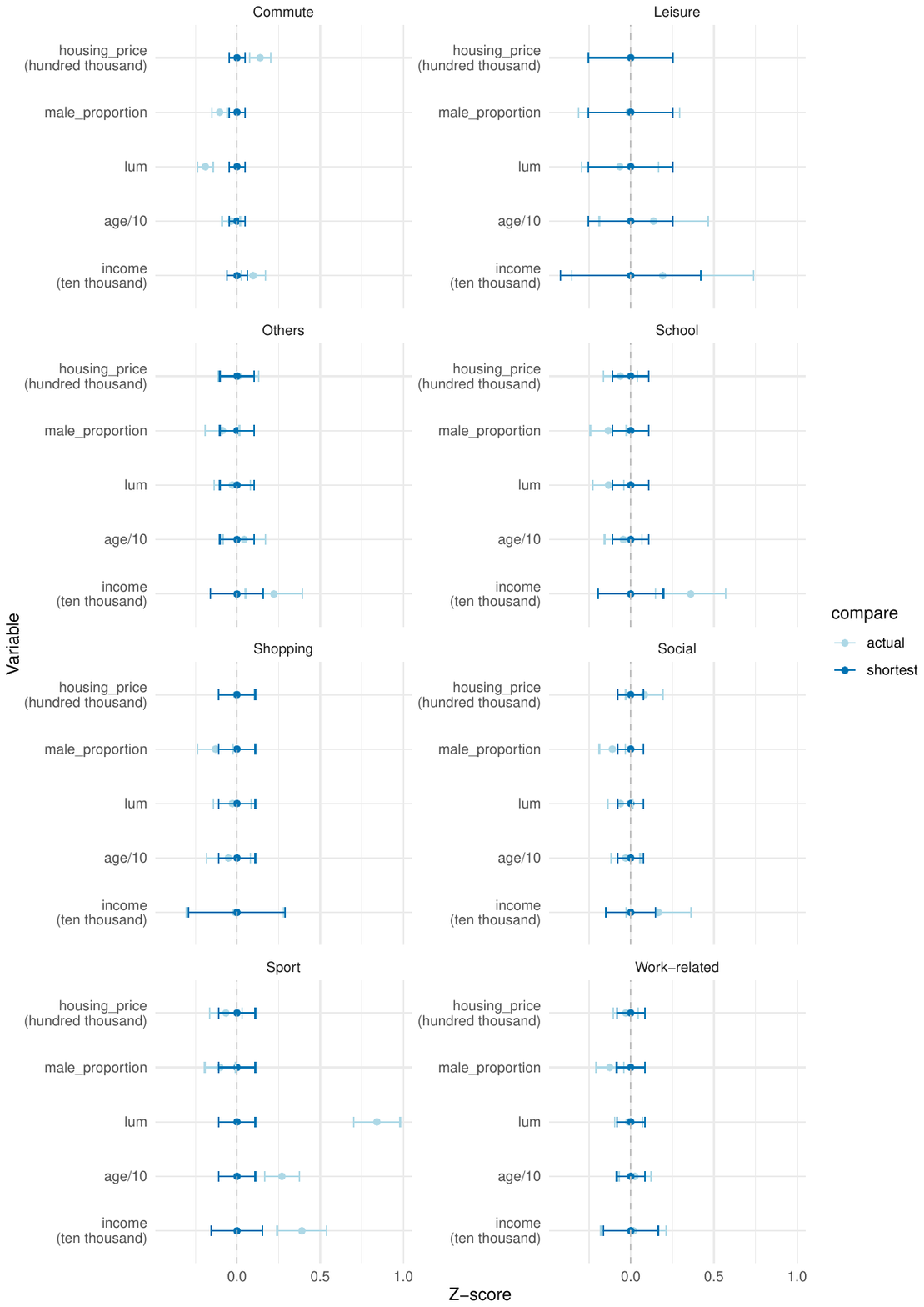}
\caption{Differences in non-visual factors between the shortest path and the actual path under different travel purposes.}
\label{fig:nonvisual_compare}
\end{figure}

\autoref{fig:nonvisual_compare} compares standardized (Z-score) values of non-visual variables between shortest and actual cyclist routes for different trip purposes. Z-scores, calculated based on shortest path values, allow meaningful comparison. Dark blue represents shortest paths, light blue actual paths, with circles showing mean values and error bars indicating 95\% confidence intervals. This analysis reveals how cyclists deviate from shortest paths based on trip purposes, potentially prioritizing certain environmental features.

Commute trips show actual paths diverging from shortest routes in key aspects. These paths traverse areas with higher housing prices, possibly reflecting preferences for safer or more aesthetically pleasing routes. However, they exhibit lower male resident proportions and less diverse land use, suggesting commuters may favor residential areas or homogeneous land use, perhaps avoiding congested commercial or industrial zones. Sports-related trips display markedly different characteristics. Chosen routes feature greater land use diversity, higher average age, and increased income levels compared to shortest paths. This may indicate preferences for varied surroundings or proximity to well-maintained recreational facilities, possibly reflecting a tendency to travel through neighborhoods perceived as conducive to outdoor activities. For other trip purposes, no statistically significant differences were found between actual and shortest paths across non-visual variables. This suggests route selection for these purposes may be influenced by more individualized factors or variables beyond this analysis's scope.

\begin{figure}[H]
\centering
\includegraphics[width=1\linewidth]{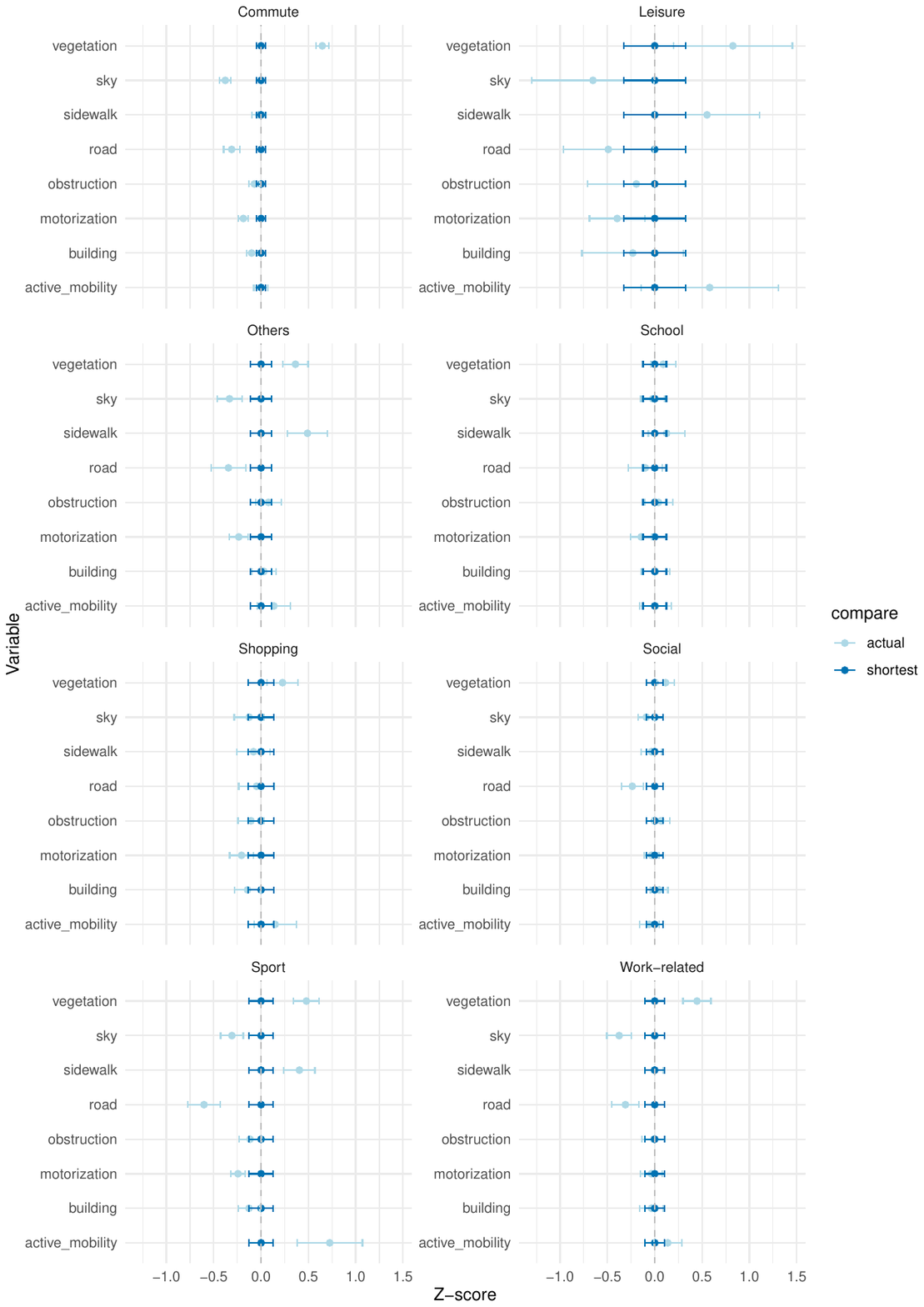}
\caption{Differences in visual factors between the shortest path and the actual path under different travel purposes.}
\label{fig:visual_compare}
\end{figure}

\autoref{fig:visual_compare} compares visual variables between shortest and actual paths. For commute and work-related trips, actual paths show higher vegetation presence and lower sky presence, suggesting preferences for greener, well-enclosed routes. These paths also have less road presence and motorization, indicating cyclists prioritize environmental quality and safety over shortest distance.
Leisure trips show higher values for vegetation, sidewalk, and active mobility, and lower values for sky, road, and motorization in actual paths. However, these differences are not statistically significant.
Social, shopping, and school trips display patterns more aligned with shortest paths, with minor preferences for lower road presence. This suggests streetscape features might be less influential for these activities.
Sport-related and other trips show strong preferences for routes with higher vegetation, sidewalks, and active mobility presence, and lower sky, road, and motorization presence. This implies a preference for safe, bike-friendly routes offering a more natural, less urbanized environment.

In summary, across all trip purposes, cyclists consistently prefer routes with better active mobility presence, less motorization, and more vegetation, even if not the shortest. This indicates safety, comfort, and environmental quality significantly influence route choice, often outweighing the desire for the shortest path.

\subsubsection{Multivariate regression analysis}
\autoref{tab:anova2} presents ANOVA test results examining factors influencing urban cycling trip purposes. This analysis uses trip purpose as the target variable and evaluates the significance of various trajectory-related variables, visual factors, and non-visual factors extracted from path deviations.

Land use mix also emerges as a highly significant factor (p < 2.90e-08), underlining the importance of urban layout diversity along the cycling routes. Among trip characteristics, distance, duration, and temperature all show high significance (p < 0.001), suggesting strong associations between journey length, environmental conditions, and trip purpose. Slope is moderately significant (p < 0.01), while speed shows only marginal significance (p < 0.07).

Socioeconomic factors present a mixed picture. After-tax income and average age both show high significance (p < 0.001), indicating that such characteristics along the deviated routes are associated with trip purposes. However, the male proportion and median housing price do not reach statistical significance.

Visual elements extracted from path deviations reveal interesting patterns. Roads, sidewalks, motorization, and active mobility all show some level of significance (p < 0.05), suggesting that the presence of these elements along deviated routes differs across trip purposes. However, buildings, vegetation, sky, and obstructions do not appear to significantly differentiate trip purposes. This implies that while specific infrastructure types may influence route choices for different trip purposes, broader landscape elements have a more uniform effect. Point of Interest (POI) density shows moderate significance (p < 0.05), indicating that destination concentration in deviated areas influences trip purposes, albeit to a lesser degree than factors such as timing and route characteristics.

These findings have important implications for urban planning and transportation policy. They suggest that strategies to promote certain types of bicycle trips should consider not only physical infrastructure but also temporal patterns, land use diversity, and socioeconomic factors. The significance of visual elements along deviated routes highlights the importance of streetscape design in influencing cycling behaviors for different trip purposes.

\begin{table}[H]
    \centering
        \caption{Non-visual and visual analysis: ANOVA test for multiple regression models based on trip purpose.}
    \resizebox{\textwidth}{!}{%
    \begin{threeparttable}
    \begin{tabular}{>{\raggedright\arraybackslash}p{4cm} >{\raggedright\arraybackslash}p{2cm} >{\raggedright\arraybackslash}p{3cm} >{\raggedright\arraybackslash}p{3cm} >{\raggedright\arraybackslash}p{3cm} >{\raggedright\arraybackslash}p{3cm}}
        \toprule
        Category & Df & Deviance & Resid. Df & Resid. Dev & Pr(>F) \\
        \midrule
        distance & 7 & 30.33 & 11235 & 3664.6 & 8.27e-05 *** \\
        temperature & 7 & 28.29 & 11235 & 3662.5 & 1.951e-04 *** \\
        duration & 7 & 39.23 & 11235 & 3673.5 & 1.76e-06 *** \\
        speed & 7 & 12.90 & 11235 & 3647.2 & 0.07 . \\
        slope & 7 & 20.31 & 11235 & 3654.6 & 4.94e-03 ** \\
        POIs & 7 & 17.72 & 11235 & 3652.0 & 0.01 * \\
        after-tax income & 7 & 36.70 & 11235 & 3671.0 & 5.35e-06 *** \\
        average age & 7 & 26.54 & 11235 & 3660.8 & 4.03e-04 *** \\
        male proportion & 7 & 11.03 & 11235 & 3645.3 & 0.14 \\
        median housing price & 7 & 10.71 & 11235 & 3645.0 & 0.15 \\
        land use mix & 7 & 48.46 & 11235 & 3682.7 & 2.90e-08 *** \\
        road & 7 & 16.07 & 11235 & 3650.3 & 0.02 * \\
        sidewalk & 7 & 14.68 & 11235 & 3648.9 & 0.04 * \\
        building & 7 & 10.91 & 11235 & 3645.2 & 0.14 \\
        vegetation & 7 & 2.83 & 11235 & 3637.1 & 0.90 \\
        sky & 7 & 3.09 & 11235 & 3637.3 & 0.88 \\
        motorization & 7 & 18.54 & 11235 & 3652.8 & 0.01 ** \\
        active mobility & 7 & 16.79 & 11235 & 3651.0 & 0.02 * \\
        obstruction & 7 & 9.86 & 11235 & 3644.1 & 0.20 \\
        \bottomrule
    \end{tabular}
    \begin{tablenotes}
        \footnotesize
        \item \textit{Significance codes:} 0 ‘***’ 0.001 ‘**’ 0.01 ‘*’ 0.05 ‘.’ 0.1 ‘ ’ 1
    \end{tablenotes}
    \end{threeparttable}
    }
    \label{tab:anova2}
\end{table}

\section{Discussions}
\label{sc:discussions}
This study investigated the effect of bicycle trip purpose on path choice, revealing significant variations across purposes, particularly for commute and sports-related trips. Commute trips favored routes with higher housing prices but lower male proportions and land use mix, while sports-related trips showed higher land use mix, average age, and income levels. These findings differ from previous studies on gender influences in cycling \citep{garrard2012women,griffin2015male,aldred2016does}. Notably, these socioeconomic characteristics might not directly influence route choice but could mediate other factors such as infrastructure quality or perceived safety.

Our results highlight disparities between actual and shortest paths, with cyclists preferring routes with more greenery, fewer vehicles, and higher pedestrian/cyclist activity, especially for recreational trips. These align with previous research on active mobility attitudes \citep{gatersleben2007contemplating,winters2011motivators,sarkar2015exploring,nawrath2019influence}. ANOVA results corroborated the significance of factors such as trip length and time of day in differentiating trip purposes.

Comparing the two models reveals interesting insights. Model 1 (non-visual analysis) yielded an AIC of 1883.177, BIC of 2387.322, and pseudo R² of 0.223 (Table \ref{tab:my_label1}, \ref{tab:my_label2}, and \ref{tab:my_label3}), while Model 2 (visual analysis) resulted in an AIC of 3970.253, BIC of 4876.631, and pseudo R² of 0.224 (Table \ref{tab:my_label_2_1}, \ref{tab:my_label_2_2}, \ref{tab:my_label_2_3}, and \ref{tab:my_label_2_4}). Despite similar pseudo R² values, Model 1's lower AIC and BIC suggest a better balance between fit and complexity, contradicting previous findings \citep{ito2021assessing}. This discrepancy likely stems from Model 2's potential overfitting due to its inclusion of both non-visual and visual features. These results underscore the critical need for robust feature selection in future bicycle route choice modeling. Researchers should focus on identifying the most influential factors from both categories, employing techniques such as regularization or machine learning-based selection algorithms to develop more parsimonious and interpretable models without sacrificing predictive power.

This study has several limitations and uncertainties that warrant acknowledgment. Primarily, our analysis relies heavily on government-sourced travel trajectory data that has undergone preliminary processing. This processing resulted in certain alterations, including intermittent data handling and obscured start- and end-points. However, we assume that this fuzzy processing has minimal impact on the acquisition of spatial information, given that the social factor data is collected at various census parcel levels. Moreover, the substantial volume of raw data (over 4,000 rows) helps mitigate the potential impact of abnormal data points, ensuring robustness in the analysis despite the imperfections in the processed data.

The relative novelty of bicycle track data acquisition services presents another limitation. Currently, we lack similar data from other cities for cross-validation purposes, highlighting the need for future research to explore broader datasets for comprehensive validation and comparison.

Data quality issues, particularly in census data and Street View Imagery (SVI) sources, also pose limitations. We utilized Google Street View (GSV) as the source of SVI, selecting sample points along travel paths at 10-meter intervals and spatially matching them to actual travel paths. Future studies could focus on identifying more accurate sample points to reduce potential bias, and they could focus on understanding the quality issues and limitations of SVI such as inconsistent weather conditions and segmentation errors~\citep{2022_jag_svi_quality}. Furthermore, it is crucial to recognize the inherent bias in SVI data perspective. As SVI is collected by vehicles, and Montreal already has an extensive bicycle network, it may not fully represent the cyclist's perspective \citep{ito_translating_2024}. The spatial parallel difference between the actual travel path and the OpenStreetMap lane demonstrates this discrepancy in spatial perspective. Addressing these limitations in future research could significantly improve the accuracy and reliability of analyses in this field.

\section{Conclusion}
\label{sc:conclusion}
We explain the distinctive characteristics of active cycling for different purposes, revealing the spatial forces that drive actual travel paths. By combining visual information on streetscapes with socioeconomic and environmental variables, we establish a new integrated framework for exposing active cycling travel patterns for the first time.

Notably, the analysis shows that cyclists choose routes with different spatial characteristics depending on the purpose of their trip: there is a significant deviation between the shortest path and the actual path. Visual factors demonstrate obvious differences in route selection. The actual path is characterized by more greenery and active mobility users, as well as less sky and motorization than the shortest path.

In essence, these findings underscore the multifaceted nature of bicycle travel behavior and highlight the importance of considering various factors --- both spatial and contextual --- to understand and improve the cycling experience. Our innovative methodology not only broadens the analytical toolkit available to researchers but also holds promise in forming urban planning and transportation policy. By investigating the underlying drivers and determinants of bicycle travel behavior, researchers can gain deeper insight into the complexities of urban mobility and pave the way for more effective and sustainable transportation strategies.

In future studies, there are further promising avenues to explore. One idea is to refine model selection and improve data resolution. This could involve exploring alternative modeling techniques, such as machine learning algorithms, to better analyze travel behavior patterns. Additionally, enhancing data collection methods, such as leveraging real-time tracking technologies, may provide more accurate and up-to-date insights into how people choose their travel paths.

Another exciting prospect is to focus on understanding individual preferences and incorporating them into navigation systems. By considering factors such as preferred routes, safety concerns, and environmental preferences, navigation systems can offer personalized route recommendations that prioritize safety and convenience for cyclists.

\section*{Acknowledgments}
The authors would like to sincerely thank the community of the Urban Analytics Lab at the National University of Singapore for helpful discussions and technical guidance on topics related to this work.
This research was funded by the Singapore International Graduate Award (SINGA) scholarship provided by the Agency for Science, Technology, and Research (A*STAR) and the NUS. 
The open data sources that made this research possible are gratefully acknowledged.
This research is part of the project Large-scale 3D Geospatial Data for Urban Analytics, which is supported by the National University of Singapore under the Start Up Grant R-295-000-171-133.

\section*{Declaration of generative AI and AI-assisted technologies in the writing process}
During the preparation of this work the authors used Claude in order to proofread the manuscript. After using this tool/service, the authors reviewed and edited the content as needed and take full responsibility for the content of the publication.

\appendix
\section*{Appendix}
The analysis in \autoref{sc:results} shows the following differential distribution of travel personal characteristics, socioeconomic elements, and environmental elements around the starting point not mentioned in the text for different trip purposes.
These are included as a supplement to provide further insights.
\begin{figure}[H]
    \begin{minipage}[t]{0.5\textwidth}
        \centering
        \includegraphics[width=\textwidth]{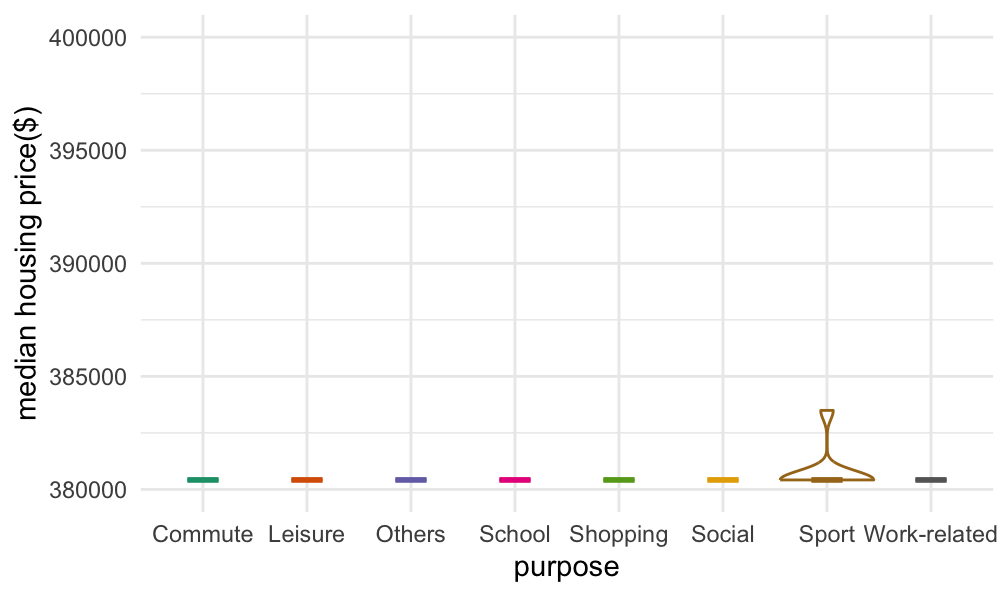}
        \caption{Median housing price distribution by cycling purpose.}
        \label{fig:housingprice}
    \end{minipage}
    \begin{minipage}[t]{0.5\textwidth}
        \centering
        \includegraphics[width=\textwidth]{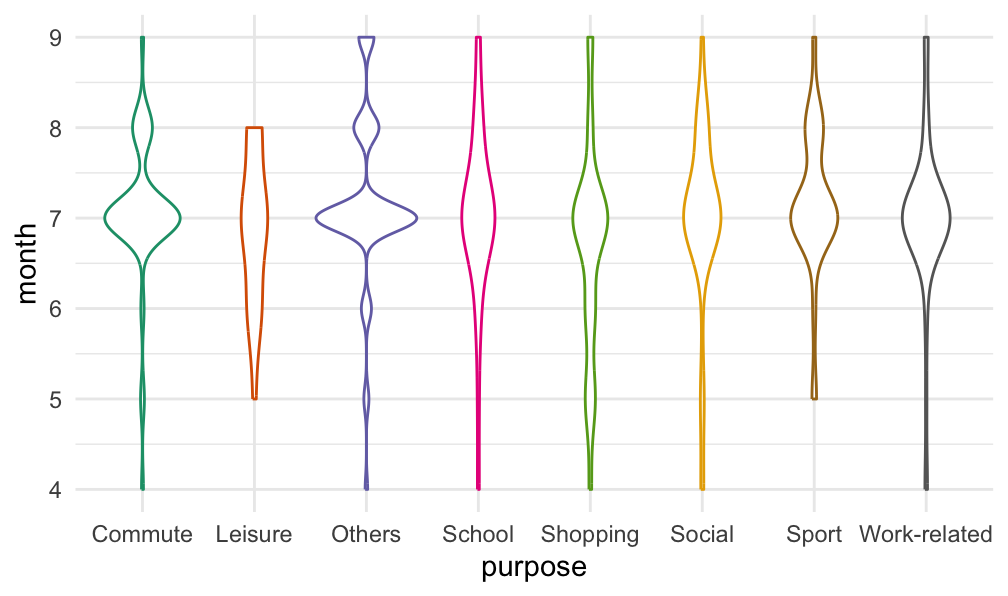}
        \caption{Travel month distribution by cycling purpose.}
        \label{fig:month}
    \end{minipage}%
\end{figure}

\begin{figure}[H]
    \begin{minipage}[t]{0.5\textwidth}
        \centering
        \includegraphics[width=\textwidth]{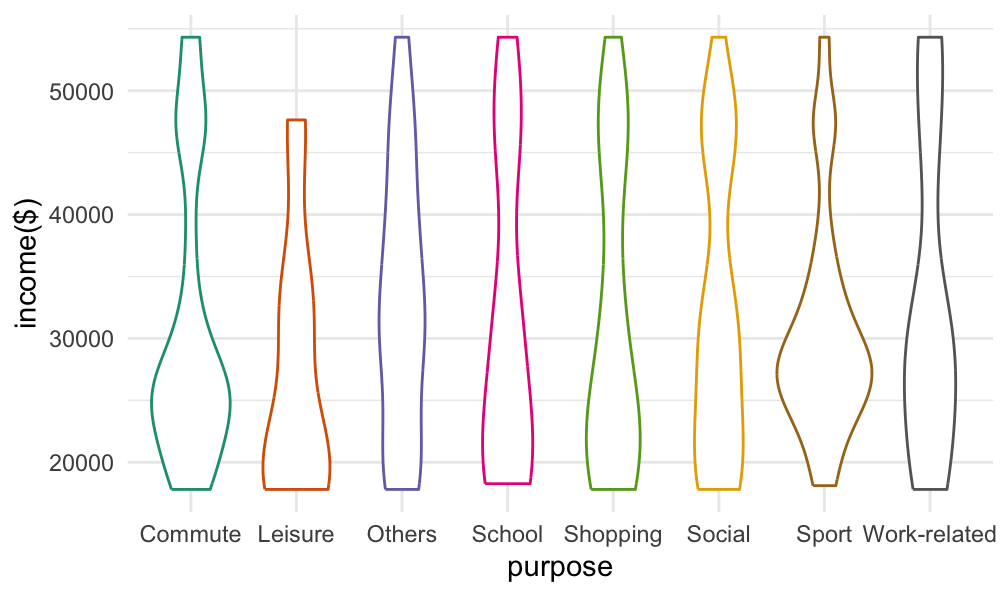}
        \caption{Income distribution by cycling purpose.}
        \label{fig:income}
    \end{minipage}
    \begin{minipage}[t]{0.5\textwidth}
        \centering
        \includegraphics[width=\textwidth]{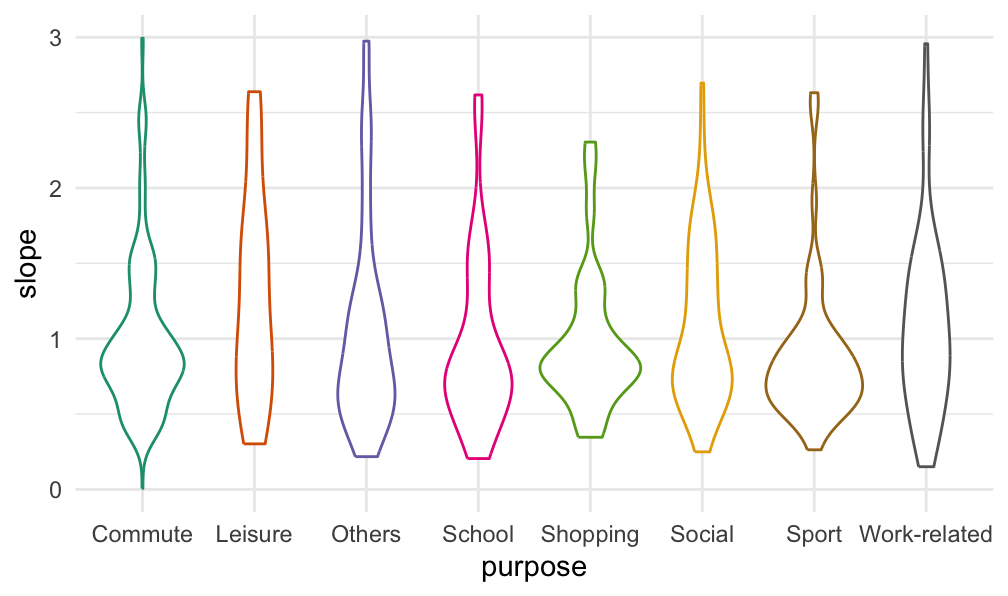}
        \caption{Mean slope distribution by cycling purpose.}
        \label{fig:slope}
    \end{minipage}%
\end{figure}

\begin{figure}[H]
    \begin{minipage}[t]{0.5\textwidth}
        \centering
        \includegraphics[width=\textwidth]{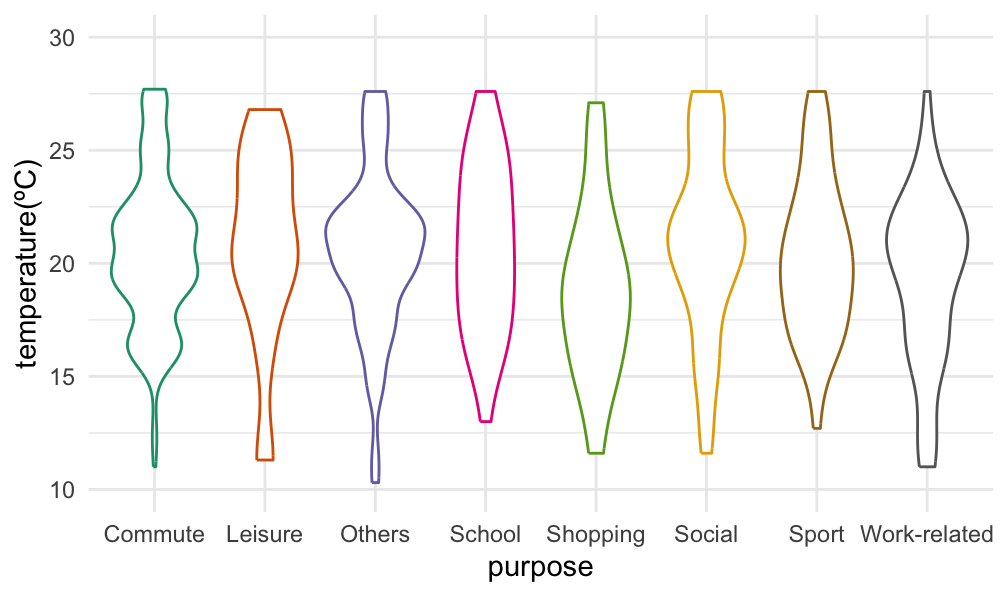}
        \caption{Temperature distribution by purpose.}
        \label{fig:temperature}
    \end{minipage}%
    \begin{minipage}[t]{0.5\textwidth}
        \centering
        \includegraphics[width=\textwidth]{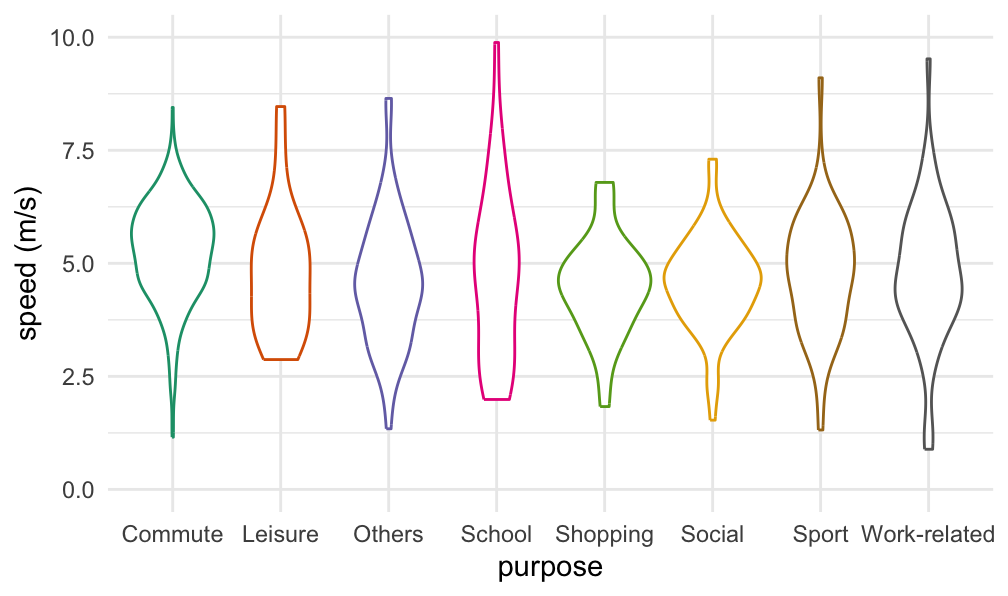}
        \caption{Speed distribution by purpose.}
        \label{fig:speed}
    \end{minipage}
\end{figure}

The following two multiple regression model summaries are taken from the models in sections 4.1.2 and 4.2.2 of the article, respectively. The trip purpose-starting point choice model is based on work-related trips, with two-by-two comparisons for other purposes to observe model regressivity (Table \ref{tab:my_label1}, \ref{tab:my_label2}, and\ref{tab:my_label3}). The regression model for the actual path-shortest path Non-Overlap section is also based on work-related
trips as a comparison against other purposes (Table \ref{tab:my_label_2_1}, \ref{tab:my_label_2_2}, \ref{tab:my_label_2_3}, and\ref{tab:my_label_2_4}).

\begin{table}[H]
    \centering
    \caption{Trip Purpose-Starting Point Choice Model Summary (1).}
    \resizebox{\textwidth}{!}{%
    \begin{threeparttable}
    \begin{tabular}{>{\raggedright\arraybackslash}p{5cm} >{\raggedright\arraybackslash}p{5cm} >{\raggedright\arraybackslash}p{5cm} >{\raggedright\arraybackslash}p{5cm} >{\raggedright\arraybackslash}p{5cm}}
        \toprule
        Category & Estimate & Std. Error & z value & Pr(>z) \\
        \midrule
        (Intercept):1 & 1.651e+01 & 6.877e+00 & NA & NA \\
        (Intercept):2 & 3.442e+01 & 1.940e+02 & 0.177 & 0.85914 \\
        (Intercept):3 & 2.757e+01 & 1.612e+01 & 1.710 & 0.08722 . \\
        (Intercept):4 & 5.973e+01 & 2.776e+02 & 0.215 & 0.82963 \\
        (Intercept):5 & 2.695e+01 & 1.881e+02 & 0.143 & 0.88604 \\
        (Intercept):6 & 2.199e+02 & 1.352e+02 & 1.626 & 0.10386 \\
        (Intercept):7 & 1.300e+02 & 1.323e+02 & 0.983 & 0.32579 \\
        length:1 & 2.678e-04 & 1.348e-04 & 1.986 & 0.04701 * \\
        length:2 & -3.081e-04 & 1.582e-04 & -1.948 & 0.05147 . \\
        length:3 & 1.669e-04 & 1.470e-04 & 1.136 & 0.25607 \\
        length:4 & 8.411e-04 & 4.525e-04 & 1.859 & 0.06308 . \\
        length:5 & -3.311e-04 & 1.954e-04 & -1.694 & 0.09024 . \\
        length:6 & 4.439e-05 & 1.618e-04 & 0.274 & 0.78380 \\
        length:7 & 1.200e-04 & 1.652e-04 & 0.726 & 0.46754 \\
        temperature:1 & 4.289e-02 & 2.573e-02 & 1.667 & 0.09549 . \\
        temperature:2 & 1.544e-02 & 3.411e-02 & 0.452 & 0.65091 \\
        temperature:3 & 3.940e-02 & 4.620e-02 & 0.853 & 0.39373 \\
        temperature:4 & 1.393e-01 & 6.225e-02 & 2.238 & 0.02521 * \\
        temperature:5 & 3.064e-02 & 4.270e-02 & 0.718 & 0.47302 \\
        temperature:6 & 6.412e-02 & 3.643e-02 & 1.760 & 0.07837 . \\
        temperature:7 & 6.831e-02 & 3.687e-02 & 1.853 & 0.06392 . \\
        duration:1 & -2.367e-02 & 3.738e-02 & -0.633 & 0.52657 \\
        duration:2 & 6.790e-02 & 4.026e-02 & 1.687 & 0.09168 . \\
        duration:3 & 2.373e-02 & 4.047e-02 & 0.586 & 0.55756 \\
        duration:4 & -3.177e-01 & 1.488e-01 & NA & NA \\
        duration:5 & 5.712e-02 & 4.890e-02 & 1.168 & 0.24282 \\
        duration:6 & 6.491e-03 & 4.283e-02 & 0.152 & 0.87954 \\
        duration:7 & -1.570e-02 & 4.468e-02 & -0.351 & 0.72522 \\
        speed:1 & 5.857e-02 & 1.455e-01 & 0.402 & 0.68734 \\
        speed:2 & 2.939e-01 & 1.528e-01 & 1.923 & 0.05444 . \\
        speed:3 & -2.134e-02 & 2.457e-01 & -0.087 & 0.93078 \\
        speed:4 & -4.904e-01 & 3.353e-01 & NA & NA \\
        speed:5 & 3.340e-01 & 1.790e-01 & 1.866 & 0.06204 . \\
        speed:6 & -1.105e-01 & 1.958e-01 & -0.565 & 0.57229 \\
        speed:7 & -1.761e-01 & 1.987e-01 & -0.886 & 0.37569 \\
        slope:1 & -2.814e-01 & 1.567e-01 & -1.796 & 0.07249 . \\
        slope:2 & -9.764e-02 & 2.194e-01 & -0.445 & 0.65625 \\
        slope:3 & -5.946e-01 & 3.841e-01 & -1.548 & 0.12158 \\
        slope:4 & -2.345e-01 & 3.030e-01 & -0.774 & 0.43911 \\
        slope:5 & -5.707e-01 & 3.674e-01 & -1.553 & 0.12035 \\
        slope:6 & -2.831e-01 & 2.326e-01 & -1.217 & 0.22354 \\
        slope:7 & -2.311e-02 & 1.841e-01 & -0.126 & 0.90007 \\
        male proportion:1 & -1.301e+01 & 6.850e+00 & -1.899 & 0.05756 . \\
        male proportion:2 & 1.904e+00 & 9.700e+00 & 0.196 & 0.84441 \\
        male proportion:3 & -3.219e+01 & 1.052e+01 & NA & NA \\
        male proportion:4 & 2.305e+00 & 1.178e+01 & 0.196 & 0.84486 \\
        male proportion:5 & -4.857e+00 & 1.092e+01 & -0.445 & 0.65650 \\
        male proportion:6 & -5.709e+00 & 9.064e+00 & -0.630 & 0.52876 \\
        male proportion:7 & -1.906e+01 & 8.859e+00 & NA & NA \\
        \bottomrule
    \end{tabular}
    \begin{tablenotes}
        \footnotesize
        \item \textit{Significance codes:} 0 ‘***’ 0.001 ‘**’ 0.01 ‘*’ 0.05 ‘.’ 0.1 ‘ ’ 1
    \end{tablenotes}
    \end{threeparttable}
    }
    \label{tab:my_label1}
\end{table}

\begin{table}[H]
    \centering
    \caption{Trip Purpose-Starting Point Choice Model Summary (2).}
    \resizebox{\textwidth}{!}{%
    \begin{threeparttable}
    \begin{tabular}{>{\raggedright\arraybackslash}p{5cm} >{\raggedright\arraybackslash}p{5cm} >{\raggedright\arraybackslash}p{5cm} >{\raggedright\arraybackslash}p{5cm} >{\raggedright\arraybackslash}p{5cm}}
        \toprule
        Category & Estimate & Std. Error & z value & Pr(>z) \\
        \midrule
        after-tax income:1 & -2.647e-07 & 1.725e-05 & -0.015 & 0.98776 \\
        after-tax income:2 & -1.560e-05 & 2.416e-05 & -0.646 & 0.51853 \\
        after-tax income:3 & 1.440e-05 & 2.893e-05 & 0.498 & 0.61871 \\
        after tax income:4 & -5.170e-05 & 3.108e-05 & -1.664 & 0.09618 . \\
        after-tax income:5 & -4.166e-05 & 2.936e-05 & -1.419 & 0.15585 \\
        after-tax income:6 & -1.818e-05 & 2.230e-05 & -0.815 & 0.41502 \\
        after-tax income:7 & 2.583e-05 & 2.267e-05 & 1.139 & 0.25464 \\
        hour:1 & 4.210e-02 & 3.565e-02 & 1.181 & 0.23766 \\
        hour:2 & 7.833e-02 & 5.523e-02 & 1.418 & 0.15614 \\
        hour:3 & 8.953e-02 & 5.186e-02 & 1.726 & 0.08429 . \\
        hour:4 & 5.348e-02 & 6.295e-02 & 0.850 & 0.39559 \\
        hour:5 & 1.496e-01 & 6.566e-02 & 2.278 & 0.02271 * \\
        hour:6 & 2.145e-01 & 4.976e-02 & 4.312 & 1.62e-05 *** \\
        hour:7 & 9.610e-02 & 4.438e-02 & 2.165 & 0.03036 * \\
        POIs:1 & 1.109e-02 & 5.366e-02 & 0.207 & 0.83634 \\
        POIs:2 & -1.747e-01 & 1.187e-01 & -1.471 & 0.14117 \\
        POIs:3 & -2.650e-01 & 1.534e-01 & -1.728 & 0.08399 . \\
        POIs:4 & 8.683e-02 & 7.194e-02 & 1.207 & 0.22748 \\
        POIs:5 & 1.401e-01 & 6.771e-02 & 2.069 & 0.03857 * \\
        POIs:6 & 1.388e-01 & 5.670e-02 & 2.448 & 0.01435 * \\
        POIs:7 & -5.363e-02 & 7.507e-02 & -0.714 & 0.47496 \\
        average age:1 & -3.284e-02 & 2.960e-02 & -1.110 & 0.26714 \\
        average age:2 & -8.924e-02 & 4.591e-02 & NA & NA \\
        average age:3 & -9.371e-02 & 4.555e-02 & -2.058 & 0.03964 * \\
        average age:4 & -2.335e-03 & 5.759e-02 & -0.041 & 0.96765 \\
        average age:5 & -2.149e-01 & 6.628e-02 & NA & NA \\
        average age:6 & -7.501e-02 & 3.790e-02 & -1.979 & 0.04779 * \\
        average age:7 & -5.011e-02 & 3.537e-02 & -1.417 & 0.15652 \\
        housing price:1 & -2.371e-05 & 1.389e-05 & -1.707 & 0.08774 . \\
        housing price:2 & -8.860e-05 & 5.096e-04 & NA & NA \\
        housing price:3 & -2.687e-05 & 3.902e-05 & NA & NA \\
        housing price:4 & -1.468e-04 & 7.298e-04 & NA & NA \\
        housing price:5 & -3.453e-05 & 4.938e-04 & NA & NA \\
        housing price:6 & -5.617e-04 & 3.555e-04 & NA & NA \\
        housing price:7 & -3.154e-04 & 3.476e-04 & NA & NA \\
        month:1 & -1.407e-01 & 1.194e-01 & -1.178 & 0.23869 \\
        month:2 & 3.100e-03 & 1.553e-01 & 0.020 & 0.98407 \\
        month:3 & 9.042e-02 & 2.109e-01 & 0.429 & 0.66813 \\
        month:4 & -4.702e-01 & 2.629e-01 & NA & NA \\
        month:5 & -4.786e-01 & 2.081e-01 & NA & NA \\
        month:6 & -2.692e-01 & 1.662e-01 & -1.620 & 0.10516 \\
        month:7 & -5.149e-02 & 1.643e-01 & -0.313 & 0.75399 \\
        day of week:1 & 1.138e+00 & 5.300e-01 & 2.148 & 0.03174 * \\
        day of week:2 & 2.971e-01 & 7.713e-01 & 0.385 & 0.70008 \\
        day of week:3 & -1.898e+00 & 5.937e-01 & -3.197 & 0.00139 ** \\
        day of week:4 & -9.898e-01 & 7.066e-01 & -1.401 & 0.16126 \\
        day of week:5 & -1.344e+00 & 6.656e-01 & -2.019 & 0.04345 * \\
        day of week:6 & -1.421e+00 & 5.435e-01 & -2.615 & 0.00892 ** \\
        day of week:7 & -8.185e-01 & 5.425e-01 & -1.509 & 0.13134 \\
        \bottomrule
    \end{tabular}
    \begin{tablenotes}
        \footnotesize
        \item \textit{Significance codes:} 0 ‘***’ 0.001 ‘**’ 0.01 ‘*’ 0.05 ‘.’ 0.1 ‘ ’ 1
    \end{tablenotes}
    \end{threeparttable}
    }
    \label{tab:my_label2}
\end{table}

\begin{table}[H]
    \centering
    \caption{Trip Purpose-Starting Point Choice Model Summary (3).}
    \resizebox{\textwidth}{!}{%
    \begin{threeparttable}
    \begin{tabular}{>{\raggedright\arraybackslash}p{5cm} >{\raggedright\arraybackslash}p{5cm} >{\raggedright\arraybackslash}p{5cm} >{\raggedright\arraybackslash}p{5cm} >{\raggedright\arraybackslash}p{5cm}}
        \toprule
        Category & Estimate & Std. Error & z value & Pr(>z) \\
        \midrule
        lum:1 & -6.455e-01 & 4.213e-01 & -1.532 & 0.12545 \\
        lum:2 & -6.856e-01 & 6.147e-01 & -1.115 & 0.26470 \\
        lum:3 & -8.456e-01 & 5.960e-01 & -1.419 & 0.15597 \\
        lum:4 & -8.644e-01 & 7.799e-01 & -1.108 & 0.26775 \\
        lum:5 & -2.571e+00 & 8.055e-01 & -3.191 & 0.00142 ** \\
        lum:6 & -9.723e-01 & 5.345e-01 & -1.819 & 0.06888 . \\
        lum:7 & -2.622e-02 & 5.013e-01 & -0.052 & 0.95828 \\
        compare path:1 & -5.167e-02 & 3.547e-02 & -1.457 & 0.14514 \\
        compare path:2 & 9.287e-03 & 3.022e-02 & 0.307 & 0.75864 \\
        compare path:3 & 2.149e-02 & 2.344e-02 & 0.917 & 0.35940 \\
        compare path:4 & 2.248e-02 & 2.348e-02 & 0.957 & 0.33837 \\
        compare path:5 & 2.224e-02 & 2.363e-02 & 0.941 & 0.34679 \\
        compare path:6 & -7.874e-02 & 1.029e-01 & -0.765 & 0.44429 \\
        compare path:7 & 1.225e-02 & 2.394e-02 & 0.512 & 0.60886 \\
        \bottomrule
    \end{tabular}
    \begin{tablenotes}
        \footnotesize
        \item \textit{Significance codes:} 0 ‘***’ 0.001 ‘**’ 0.01 ‘*’ 0.05 ‘.’ 0.1 ‘ ’ 1
        \item Residual deviance: 1659.177 on 4550 degrees of freedom
        \item Log-likelihood: -829.5886 on 4550 degrees of freedom
        \item AIC: 1883.177
        \item BIC: 2387.322
        \item Pseudo $R^2$: 0.223
        \item commute(1), school(2), sport (3), leisure(4), shopping(5), social(6), others(7), work-related(base)
    \end{tablenotes}
    \end{threeparttable}
    }
    \label{tab:my_label3}
\end{table}

\begin{table}[H]
    \centering
    \caption{Summary for non-overlapping route choosing model (1).}
    \resizebox{\textwidth}{!}{%
    \begin{threeparttable}
    \begin{tabular}{>{\raggedright\arraybackslash}p{5cm} >{\raggedright\arraybackslash}p{5cm} >{\raggedright\arraybackslash}p{5cm} >{\raggedright\arraybackslash}p{5cm} >{\raggedright\arraybackslash}p{5cm}}
        \toprule
        Category & Estimate & Std. Error & z value & Pr(>z) \\
        \midrule
        (Intercept):1 & -4.520e+00 & 7.824e+00 & -0.578 & 0.563428 \\
        (Intercept):2 & -5.938e+00 & 1.383e+01 & NA & NA \\
        (Intercept):3 & -1.580e+01 & 1.007e+01 & NA & NA \\
        (Intercept):4 & 2.135e+01 & 2.110e+01 & 1.012 & 0.311451 \\
        (Intercept):5 & -2.622e+01 & 1.748e+01 & NA & NA \\
        (Intercept):6 & -1.113e+01 & 9.795e+00 & NA & NA \\
        (Intercept):7 & -6.187e+00 & 1.007e+01 & NA & NA \\
        distance:1 & 1.018e-04 & 5.029e-05 & 2.025 & 0.042881 * \\
        distance:2 & 8.231e-05 & 6.548e-05 & 1.257 & 0.208772 \\
        distance:3 & 3.501e-05 & 5.675e-05 & 0.617 & 0.537264 \\
        distance:4 & -6.945e-04 & 2.269e-04 & NA & NA \\
        distance:5 & 7.899e-05 & 7.382e-05 & 1.070 & 0.284573 \\
        distance:6 & 1.317e-04 & 6.063e-05 & 2.171 & 0.029899 * \\
        distance:7 & 1.604e-04 & 5.559e-05 & 2.885 & 0.003915 ** \\
        temperature:1 & -2.114e-02 & 1.851e-02 & -1.142 & 0.253342 \\
        temperature:2 & -7.305e-02 & 2.248e-02 & -3.250 & 0.001155 ** \\
        temperature:3 & 5.224e-02 & 2.886e-02 & 1.810 & 0.070304 . \\
        temperature:4 & 1.237e-01 & 7.758e-02 & 1.595 & 0.110714 \\
        temperature:5 & -1.628e-03 & 3.308e-02 & -0.049 & 0.960737 \\
        temperature:6 & -1.454e-02 & 2.226e-02 & -0.653 & 0.513703 \\
        temperature:7 & -1.362e-02 & 2.378e-02 & -0.573 & 0.566692 \\
        duration:1 & -1.078e-02 & 7.299e-03 & -1.477 & 0.139788 \\
        duration:2 & 2.635e-04 & 9.903e-03 & 0.027 & 0.978770 \\
        duration:3 & 1.411e-02 & 7.173e-03 & 1.968 & 0.049102 * \\
        duration:4 & -1.319e-02 & 2.430e-02 & -0.543 & 0.587302 \\
        duration:5 & 3.782e-04 & 1.157e-02 & 0.033 & 0.973931 \\
        duration:6 & 3.621e-03 & 8.484e-03 & 0.427 & 0.669504 \\
        duration:7 & 1.461e-02 & 7.273e-03 & 2.009 & 0.044578 * \\
        speed:1 & 1.497e-01 & 1.010e-01 & 1.483 & 0.138171 \\
        speed:2 & -9.538e-02 & 1.448e-01 & -0.659 & 0.510118 \\
        speed:3 & 2.225e-01 & 1.140e-01 & 1.952 & 0.050981 . \\
        speed:4 & -1.849e-01 & 2.889e-01 & -0.640 & 0.522077 \\
        speed:5 & 1.004e-01 & 1.537e-01 & 0.653 & 0.513610 \\
        speed:6 & 3.182e-02 & 1.218e-01 & 0.261 & 0.793918 \\
        speed:7 & -2.390e-02 & 1.189e-01 & -0.201 & 0.840658 \\
        slope:1 & -1.940e-01 & 1.104e-01 & -1.757 & 0.078920 . \\
        slope:2 & -2.597e-01 & 2.285e-01 & -1.136 & 0.255814 \\
        slope:3 & 4.881e-02 & 1.173e-01 & 0.416 & 0.677272 \\
        slope:4 & -8.492e-01 & 5.551e-01 & NA & NA \\
        slope:5 & 1.275e-01 & 1.318e-01 & 0.968 & 0.333286 \\
        slope:6 & -1.265e-01 & 1.608e-01 & -0.787 & 0.431450 \\
        slope:7 & 1.305e-01 & 1.166e-01 & 1.119 & 0.263254 \\
        start hour:1 & 5.749e-03 & 2.326e-02 & 0.247 & 0.804760 \\
        start hour:2 & 1.095e-02 & 3.541e-02 & 0.309 & 0.757019 \\
        start hour:3 & 8.942e-02 & 2.990e-02 & 2.991 & 0.002782 ** \\
        start hour:4 & -5.503e-02 & 7.673e-02 & -0.717 & 0.473200 \\
        start hour:5 & 1.082e-01 & 4.091e-02 & 2.644 & 0.008182 ** \\
        start hour:6 & 1.960e-01 & 3.057e-02 & 6.410 & 1.45e-10 *** \\
        start hour:7 & 8.215e-02 & 2.938e-02 & 2.796 & 0.005170 ** \\
        \bottomrule
    \end{tabular}
    \begin{tablenotes}
        \footnotesize
        \item \textit{Significance codes:} 0 ‘’ 0.001 ‘’ 0.01 ‘’ 0.05 ‘.’ 0.1 ‘ ’ 1
    \end{tablenotes}
    \end{threeparttable}
    }
    \label{tab:my_label_2_1}
\end{table}

\begin{table}[H]
    \centering
    \caption{Summary for non-overlapping route choosing model (2).}
    \resizebox{\textwidth}{!}{%
    \begin{threeparttable}
    \begin{tabular}{>{\raggedright\arraybackslash}p{5cm} >{\raggedright\arraybackslash}p{5cm} >{\raggedright\arraybackslash}p{5cm} >{\raggedright\arraybackslash}p{5cm} >{\raggedright\arraybackslash}p{5cm}}
        \toprule
        Category & Estimate & Std. Error & z value & Pr(>z) \\
        \midrule
        month:1 & -3.008e-02 & 9.008e-02 & -0.334 & 0.738424 \\
        month:2 & 7.363e-02 & 1.077e-01 & 0.684 & 0.494110 \\
        month:3 & -1.439e-01 & 1.423e-01 & -1.011 & 0.312111 \\
        month:4 & -9.611e-01 & 3.864e-01 & NA & NA \\
        month:5 & -1.560e-01 & 1.574e-01 & -0.991 & 0.321454 \\
        month:6 & 7.320e-02 & 1.074e-01 & 0.681 & 0.495626 \\
        month:7 & 1.289e-01 & 1.158e-01 & 1.114 & 0.265487 \\
        day of week:1 & 1.210e+00 & 3.563e-01 & 3.397 & 0.000682 *** \\
        day of week:2 & 4.651e-01 & 5.655e-01 & 0.823 & 0.410787 \\
        day of week:3 & -1.075e+00 & 3.735e-01 & -2.879 & 0.003988 ** \\
        day of week:4 & -1.375e+00 & 7.874e-01 & -1.747 & 0.080644 . \\
        day of week:5 & -1.052e+00 & 4.764e-01 & -2.209 & 0.027199 * \\
        day of week:6 & -1.591e+00 & 3.591e-01 & -4.431 & 9.37e-06 *** \\
        day of week:7 & -9.620e-01 & 3.739e-01 & -2.573 & 0.010077 * \\
        compare path:1 & -6.640e-01 & 9.738e-02 & -6.818 & 9.20e-12 *** \\
        compare path:2 & -1.200e-02 & 3.459e-02 & -0.347 & 0.728739 \\
        compare path:3 & 6.111e-02 & 2.958e-02 & 2.066 & 0.038850 * \\
        compare path:4 & 8.820e-02 & 6.225e-02 & 1.417 & 0.156556 \\
        compare path:5 & 1.787e-02 & 5.021e-02 & 0.356 & 0.721906 \\
        compare path:6 & -3.446e-01 & 9.890e-02 & -3.484 & 0.000493 *** \\
        compare path:7 & -2.701e-02 & 3.606e-02 & -0.749 & 0.453817 \\
        POIs:1 & 9.277e-02 & 5.149e-02 & 1.802 & 0.071623 . \\
        POIs:2 & 5.630e-02 & 6.917e-02 & 0.814 & 0.415664 \\
        POIs:3 & 1.127e-01 & 6.380e-02 & 1.766 & 0.077365 . \\
        POIs:4 & 1.306e-01 & 1.108e-01 & 1.179 & 0.238222 \\
        POIs:5 & 2.083e-01 & 6.124e-02 & 3.401 & 0.000670 *** \\
        POIs:6 & 1.473e-01 & 5.704e-02 & 2.582 & 0.009825 ** \\
        POIs:7 & 4.949e-02 & 6.569e-02 & 0.753 & 0.451228 \\
        after-tax income:1 & -2.916e-05 & 1.385e-05 & -2.105 & 0.035291 * \\
        after-tax income:2 & 3.679e-05 & 2.008e-05 & 1.832 & 0.066905 . \\
        after-tax income:3 & 3.190e-06 & 1.817e-05 & 0.176 & 0.860666 \\
        after-tax income:4 & 2.258e-05 & 4.303e-05 & 0.525 & 0.599715 \\
        after-tax income:5 & -1.181e-05 & 2.251e-05 & -0.525 & 0.599844 \\
        after-tax income:6 & 1.040e-05 & 1.712e-05 & 0.607 & 0.543537 \\
        after-tax income:7 & 3.015e-05 & 1.783e-05 & 1.691 & 0.090850 . \\
        average age:1 & 8.575e-02 & 5.243e-02 & 1.635 & 0.101948 \\
        average age:2 & -1.596e-01 & 7.956e-02 & NA & NA \\
        average age:3 & 2.020e-01 & 6.541e-02 & 3.088 & 0.002015 ** \\
        average age:4 & 9.552e-02 & 1.340e-01 & 0.713 & 0.476074 \\
        average age:5 & 1.941e-02 & 8.676e-02 & 0.224 & 0.822969 \\
        average age:6 & 5.242e-02 & 6.459e-02 & 0.811 & 0.417086 \\
        average age:7 & 1.133e-02 & 6.487e-02 & 0.175 & 0.861333 \\
        male proportion:1 & -9.040e+00 & 5.015e+00 & -1.803 & 0.071460 . \\
        male proportion:2 & -1.255e+01 & 8.306e+00 & NA & NA \\
        male proportion:3 & 1.573e+00 & 6.267e+00 & 0.251 & 0.801816 \\
        male proportion:4 & -1.542e+01 & 1.476e+01 & NA & NA \\
        male proportion:5 & -7.135e+00 & 9.070e+00 & NA & NA \\
        male proportion:6 & -9.089e-01 & 6.334e+00 & -0.143 & 0.885901 \\
        male proportion:7 & 2.620e-01 & 6.173e+00 & 0.042 & 0.966148 \\
        \bottomrule
    \end{tabular}
    \begin{tablenotes}
        \footnotesize
        \item \textit{Significance codes:} 0 ‘***’ 0.001 ‘**’ 0.01 ‘*’ 0.05 ‘.’ 0.1 ‘ ’ 1
    \end{tablenotes}
    \end{threeparttable}
    }
    \label{tab:my_label_2_2}
\end{table}

\begin{table}[H]
    \centering
    \caption{Summary for non-overlapping route choosing model (3).}
    \resizebox{\textwidth}{!}{%
    \begin{threeparttable}
    \begin{tabular}{>{\raggedright\arraybackslash}p{5cm} >{\raggedright\arraybackslash}p{5cm} >{\raggedright\arraybackslash}p{5cm} >{\raggedright\arraybackslash}p{5cm} >{\raggedright\arraybackslash}p{5cm}}
        \toprule
        Category & Estimate & Std. Error & z value & Pr(>z) \\
        \midrule
        median housing price:1 & -3.462e-06 & 1.594e-06 & -2.172 & 0.029872 * \\
        median housing price:2 & -9.180e-06 & 5.076e-06 & NA & NA \\
        median housing price:3 & -1.089e-07 & 2.037e-06 & -0.053 & 0.957354 \\
        median housing price:4 & -8.638e-06 & 7.279e-06 & NA & NA \\
        median housing price:5 & -2.520e-06 & 3.126e-06 & -0.806 & 0.420159 \\
        median housing price:6 & -2.970e-06 & 2.182e-06 & -1.361 & 0.173471 \\
        median housing price:7 & -1.749e-06 & 2.173e-06 & -0.805 & 0.420853 \\
        land use mix:1 & 1.380e+00 & 3.049e-01 & 4.524 & 6.05e-06 *** \\
        land use mix:2 & 1.163e+00 & 4.592e-01 & 2.533 & 0.011318 * \\
        land use mix:3 & 1.173e+00 & 3.575e-01 & 3.281 & 0.001035 ** \\
        land use mix:4 & -8.639e-01 & 1.009e+00 & -0.856 & 0.391963 \\
        land use mix:5 & 6.009e-01 & 5.162e-01 & 1.164 & 0.244376 \\
        land use mix:6 & 6.723e-02 & 3.758e-01 & 0.179 & 0.858018 \\
        land use mix:7 & 4.203e-01 & 3.624e-01 & 1.160 & 0.246108 \\
        road:1 & 1.155e+01 & 6.151e+00 & 1.878 & 0.060352 . \\
        road:2 & 3.003e+01 & 1.265e+01 & 2.374 & 0.017616 * \\
        road:3 & 2.739e+00 & 7.316e+00 & 0.374 & 0.708081 \\
        road:4 & -1.773e+01 & 1.340e+01 & NA & NA \\
        road:5 & 2.946e+01 & 1.680e+01 & 1.754 & 0.079429 . \\
        road:6 & 5.406e+00 & 7.570e+00 & 0.714 & 0.475130 \\
        road:7 & 1.257e+00 & 7.575e+00 & 0.166 & 0.868207 \\
        sidewalk:1 & 4.414e+01 & 2.341e+01 & 1.885 & 0.059377 . \\
        sidewalk:2 & 3.302e+01 & 3.580e+01 & 0.922 & 0.356363 \\
        sidewalk:3 & 6.011e+01 & 2.809e+01 & 2.140 & 0.032377 * \\
        sidewalk:4 & 5.186e+01 & 6.165e+01 & 0.841 & 0.400263 \\
        sidewalk:5 & 1.098e+02 & 3.444e+01 & 3.188 & 0.001433 ** \\
        sidewalk:6 & 1.891e+01 & 2.871e+01 & 0.659 & 0.510176 \\
        sidewalk:7 & 6.061e+01 & 2.686e+01 & 2.256 & 0.024065 * \\
        building:1 & -1.236e+01 & 1.068e+01 & -1.157 & 0.247398 \\
        building:2 & -5.420e+00 & 2.459e+01 & -0.220 & 0.825541 \\
        building:3 & 1.734e+01 & 1.632e+01 & 1.062 & 0.288185 \\
        building:4 & 1.818e+01 & 2.877e+01 & 0.632 & 0.527359 \\
        building:5 & 2.805e+01 & 3.430e+01 & 0.818 & 0.413603 \\
        building:6 & 1.619e+01 & 1.578e+01 & 1.026 & 0.304946 \\
        building:7 & 4.716e+00 & 1.423e+01 & 0.331 & 0.740383 \\
        vegetation:1 & 7.039e+00 & 8.419e+00 & 0.836 & 0.403124 \\
        vegetation:2 & 1.706e+01 & 1.387e+01 & 1.230 & 0.218720 \\
        vegetation:3 & 7.983e+00 & 1.137e+01 & 0.702 & 0.482788 \\
        vegetation:4 & 1.203e+01 & 2.557e+01 & 0.470 & 0.638069 \\
        vegetation:5 & 2.282e+01 & 1.869e+01 & 1.221 & 0.222126 \\
        vegetation:6 & 1.127e+01 & 1.052e+01 & 1.071 & 0.284300 \\
        vegetation:7 & 9.424e+00 & 1.183e+01 & 0.797 & 0.425562 \\
        sky:1 & 3.515e+00 & 9.481e+00 & 0.371 & 0.710849 \\
        sky:2 & 1.086e+01 & 1.520e+01 & 0.715 & 0.474880 \\
        sky:3 & 3.471e+00 & 1.299e+01 & 0.267 & 0.789367 \\
        sky:4 & -1.577e+00 & 2.908e+01 & -0.054 & 0.956758 \\
        sky:5 & 2.863e+01 & 2.014e+01 & 1.422 & 0.155109 \\
        sky:6 & 1.077e+01 & 1.189e+01 & 0.906 & 0.364881 \\
        sky:7 & 3.962e+00 & 1.325e+01 & 0.299 & 0.764884 \\
        \bottomrule
    \end{tabular}
    \begin{tablenotes}
        \footnotesize
        \item \textit{Significance codes:} 0 ‘***’ 0.001 ‘**’ 0.01 ‘*’ 0.05 ‘.’ 0.1 ‘ ’ 1
    \end{tablenotes}
    \end{threeparttable}
    }
    \label{tab:my_label_2_3}
\end{table}

\begin{table}[H]
    \centering
    \caption{Summary for non-overlapping route choosing model (4).}
    \resizebox{\textwidth}{!}{%
    \begin{threeparttable}
    \begin{tabular}{>{\raggedright\arraybackslash}p{5cm} >{\raggedright\arraybackslash}p{5cm} >{\raggedright\arraybackslash}p{5cm} >{\raggedright\arraybackslash}p{5cm} >{\raggedright\arraybackslash}p{5cm}}
        \toprule
        Category & Estimate & Std. Error & z value & Pr(>z) \\
        \midrule
        motorization:1 & 8.822e+00 & 7.490e+00 & 1.178 & 0.238843 \\
        motorization:2 & 2.348e+01 & 1.334e+01 & 1.760 & 0.078382 . \\
        motorization:3 & -8.326e+00 & 1.020e+01 & -0.816 & 0.414356 \\
        motorization:4 & -2.271e+01 & 2.044e+01 & -1.111 & 0.266608 \\
        motorization:5 & 2.789e+01 & 1.787e+01 & 1.561 & 0.118603 \\
        motorization:6 & 4.993e+00 & 9.295e+00 & 0.537 & 0.591179 \\
        motorization:7 & -1.778e+01 & 1.038e+01 & -1.713 & 0.086710 . \\
        active mobility:1 & 3.896e+00 & 1.222e+01 & 0.319 & 0.749844 \\
        active mobility:2 & 3.637e+01 & 2.004e+01 & 1.815 & 0.069483 . \\
        active mobility:3 & 1.093e+01 & 1.433e+01 & 0.763 & 0.445586 \\
        active mobility:4 & -5.775e+01 & 3.891e+01 & -1.484 & 0.137790 \\
        active mobility:5 & 4.001e+01 & 2.579e+01 & 1.551 & 0.120805 \\
        active mobility:6 & 1.430e+01 & 1.454e+01 & 0.984 & 0.325345 \\
        active mobility:7 & -2.545e+01 & 1.657e+01 & -1.535 & 0.124661 \\
        obstruction:1 & 1.419e+01 & 1.062e+01 & 1.336 & 0.181418 \\
        obstruction:2 & 2.092e+01 & 2.667e+01 & 0.784 & 0.432804 \\
        obstruction:3 & -1.639e+01 & 1.525e+01 & -1.075 & 0.282346 \\
        obstruction:4 & -1.631e+01 & 2.438e+01 & -0.669 & 0.503564 \\
        obstruction:5 & -2.736e+00 & 3.841e+01 & -0.071 & 0.943228 \\
        obstruction:6 & -7.489e+00 & 1.578e+01 & -0.475 & 0.634981 \\
        obstruction:7 & -1.089e+00 & 1.295e+01 & -0.084 & 0.933009 \\
        \bottomrule
    \end{tabular}
    \begin{tablenotes}
        \footnotesize
        \item \textit{Significance codes:} 0 ‘***’ 0.001 ‘**’ 0.01 ‘*’ 0.05 ‘.’ 0.1 ‘ ’ 1
        \item Residual deviance: 3634.253 on 11228 degrees of freedom
        \item Log-likelihood: -1817.127 on 11228 degrees of freedom
        \item AIC: 3970.253
        \item BIC: 4876.631
        \item Pseudo $R^2$: 0.224
        \item commute(1), school(2), sport (3), leisure(4), shopping(5), social(6), others(7), work-related(base)
    \end{tablenotes}
    \end{threeparttable}
    }
    \label{tab:my_label_2_4}
\end{table}

\bibliographystyle{cas-model2-names}
\bibliography{reference, koichi}

\end{document}